\documentclass{article}
\usepackage{amsmath}
\usepackage{amssymb}
\usepackage{graphicx}
\usepackage{booktabs}
\usepackage{multirow}
\usepackage{graphicx, wrapfig}
\usepackage[table]{xcolor}
\definecolor{decoderblue}{RGB}{226,238,255}
\definecolor{encoderorange}{RGB}{255,239,219}
\definecolor{oursgreen}{RGB}{226,246,226}
\usepackage{algorithm}
\usepackage{algpseudocode}
\usepackage{subcaption}

\usepackage[preprint]{neurips_2026}

\usepackage[utf8]{inputenc} % allow utf-8 input
\usepackage[T1]{fontenc}    % use 8-bit T1 fonts
\usepackage{hyperref}       % hyperlinks
\usepackage{url}            % simple URL typesetting
\usepackage{booktabs}       % professional-quality tables
\usepackage{amsfonts}       % blackboard math symbols
\usepackage{nicefrac}       % compact symbols for 1/2, etc.
\usepackage{microtype}      % microtypography
\usepackage{xcolor}         % colors
\usepackage{wrapfig}

\newcommand{\tool}{\textsc{EpiNarrate}}

\title{EpiNarrate: Agentic Generation of Grounded Narratives from Epidemiological Scenario Projections}
\author{%
  Rituparna Datta \\
  Department of Computer Science\\
  University of Virginia\\
  \And
  Srini Venkatramanan \\
  Biocomplexity Institute\\
  University of Virginia\\
  \And
  Bryan L. Lewis \\
  Biocomplexity Institute\\
  University of Virginia\\
  \And
  Yiqi Su \\
  Department of Computer Science\\
  Virginia Tech\\
  \And
  Harry Hochheiser \\
  Department of Biomedical Informatics\\
  University of Pittsburgh\\
  \And
  Lucie Contamin \\
  Department of Biomedical Informatics\\
  University of Pittsburgh\\
  \And
  Parantapa Bhattacharya \\
  Biocomplexity Institute\\
  University of Virginia\\
  \And
  Naren Ramakrishnan \\
  Department of Computer Science\\
  Virginia Tech\\
  \And
  Anil Vullikanti \\
  Biocomplexity Institute and\\Department of Computer Science\\
  University of Virginia\\
}

\begin{document}

\maketitle

\begin{abstract}
Generation of clear and accessible public health narratives is critical for communicating complex epidemiological projections to policymakers and the general public at large. Such narratives require more than simply reporting numbers: projections must be contextualized and quantitatively grounded across multiple dimensions. 
Further, projections are often derived from large ensemble datasets which combine intervention assumptions, geographic and demographic strata, outcomes, time horizons, and uncertainty quantiles. 
However, directly using large language models (LLMs) to summarize and contextualize such data often leads to inconsistencies, omissions, and fragile behavior.
We introduce \tool{}, an agentic framework for public health report generation that separates structured numerical reasoning from natural-language generation. 
The framework first extracts scenario axes and organizes them into a partial-order schema, enabling systematic traversal of the underlying multidimensional space. It then constructs an augmented dataset and derives valid quantitative statements through a comparison grammar that enforces semantic and arithmetic consistency.
To balance coverage and non-redundancy, we introduce an interestingness-driven selection mechanism based on maximum-entropy principles.
Experiments on the COVID-19 Scenario Modeling Hub demonstrate that \tool{} produces narratives with improved factual grounding and broader coverage of salient epidemiological patterns, while preserving the style of expert-written reports.
\end{abstract}
\section{Introduction}
\label{sec:intro}

\begin{wrapfigure}[32]{r}{3in}
    \centering
    % \vspace{-2em}
     \begin{subfigure}{0.95\linewidth}
        \centering
        \includegraphics[width=\linewidth]{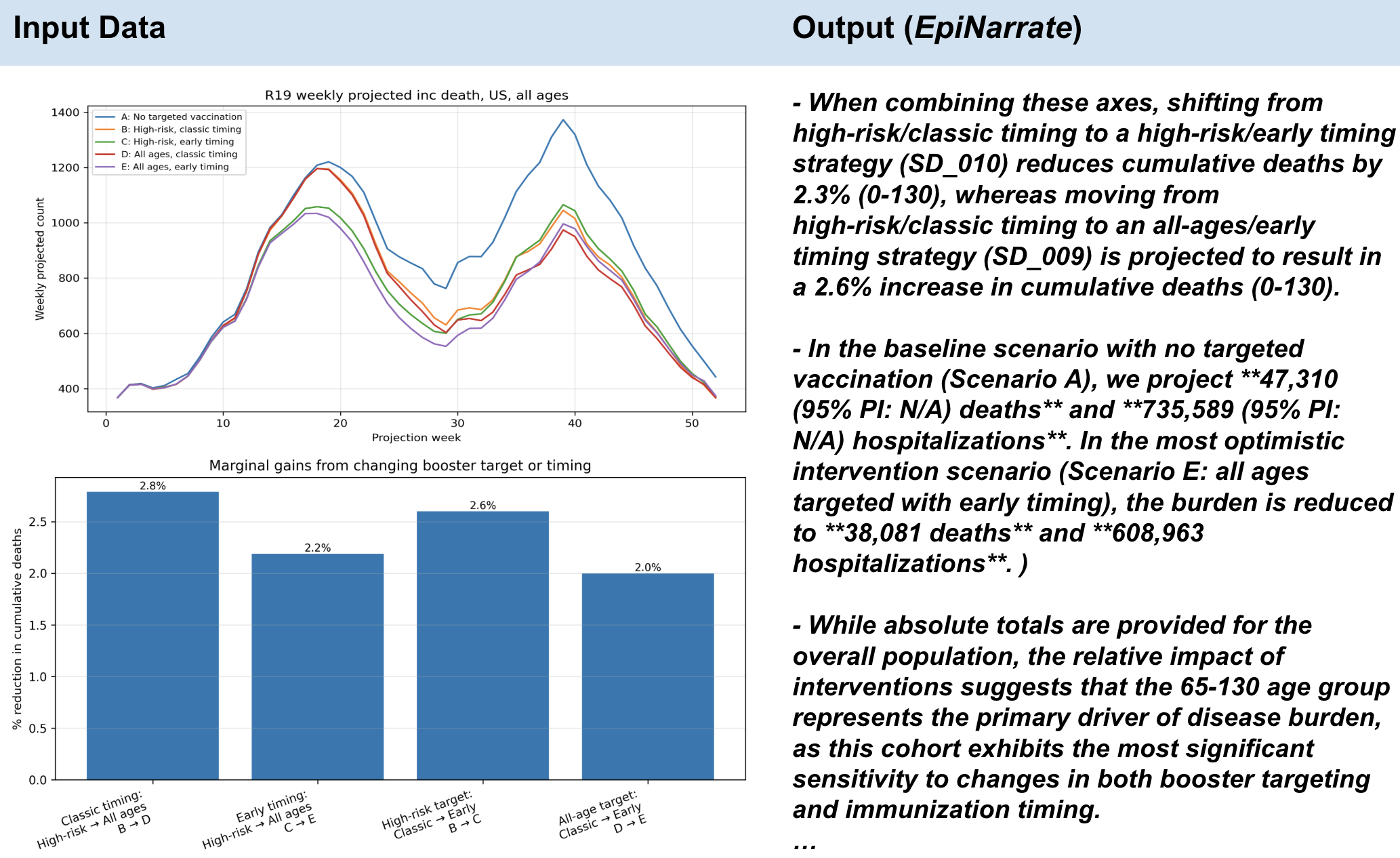}
        \caption{Example input-output narrative}
        \label{fig:dataflow_b}
    \end{subfigure}
    
    \begin{subfigure}{0.75\linewidth}
        \centering
        \includegraphics[height=2.8cm]{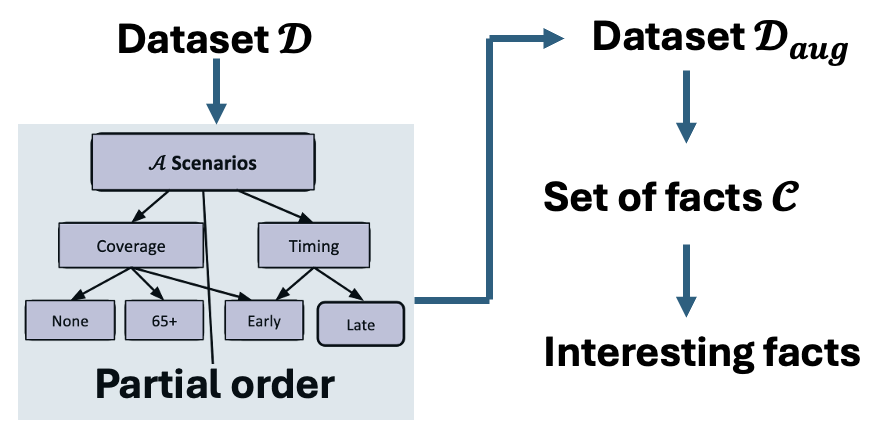}
        \caption{Data flow Abstraction}
        \label{fig:dataflow_a}
    \end{subfigure}

    \caption{
    \footnotesize
    (a) Example input and output from \tool{} for an SMH round. The input consists of structured projection data, while the output is a grounded narrative that reports scenario contrasts, burden, intervention effects, and age-specific interpretation.  (b) Data flow in \tool{} has the following structure: \textit{Prompt} $\;\leftrightarrow\;$ \textit{Partial Order} $\;\rightarrow\;$ \textit{Augmented dataset} $\;\rightarrow\;$ \textit{Set of facts} $\;\rightarrow\;$ \textit{Interesting facts}. The partial order is an abstract representation of the relationship between attribute values, and supports easier verification. It is used to construct an augmented dataset and then a set of facts, using grammars. Interesting facts are generated using an entropy based approach. The components here map to those in the architecture of \tool{}, shown in Figure \ref{fig:workflow}}
    \label{fig:dataflow}
    \vspace{-2em}
\end{wrapfigure}

Public health agencies~\cite{cdc-mmwr, cdc-smh-summary, cdc-flusight, who} rely on carefully
crafted, evidence-backed, statements (termed narratives) to communicate key information to
the broader public. 
These can include information about ongoing and forecasted trends, e.g., Flu \cite{cdc-flusight} or COVID-19 incidence
projections~\cite{cdc-smh-summary}, and
are typically written by
domain experts to contextualize results
from models or ensembles.
For instance, the COVID-19 Scenario Modeling Hub (SMH) constructs multi-model ensembles to provide robust, long-term projections to guide federal and state-level responses~\cite{midas-covid19-scenario-modeling-hub, reich2022collaborative}.
Summary reports of these ensembles are produced \cite{cdc-smh-summary}, and are used by epidemiologists as well as the lay public.
Figure \ref{fig:dataflow}(a) shows an example of such a summary.

Summary generation, as we study here, is a canonical task for LLMs and is closely related
to the broader theme of data-to-text generation.
Prior work has studied this problem in settings such as table-to-text generation, where highlighted cells from Wikipedia tables are turned into faithful textual descriptions \citep{parikh2020totto}, and sports-summary generation, where systems first select relevant records and plan their ordering before generating a coherent narrative \citep{puduppully2019data}. Other work has compared modular pipeline systems with end-to-end neural generation, showing that explicit intermediate representations can improve generalization and output quality \citep{ferreira2019neural}. However, public-health scenario-modeling reports differ from these standard settings because their inputs include large, multidimensional ensemble projections, scenario assumptions, geographic and demographic strata, and uncertainty quantiles. Producing consumable reports
requires not just summarizing the given data but placing them in context, grounding comparisons, and highlighting key takeaways. 
The motivating question for our work is thus: \emph{can agentic systems be developed to prepare narratives for public health information?}
% We refer to this as the \prob{} problem.

\begin{figure*}[h]
    \centering
    \includegraphics[width=\linewidth]{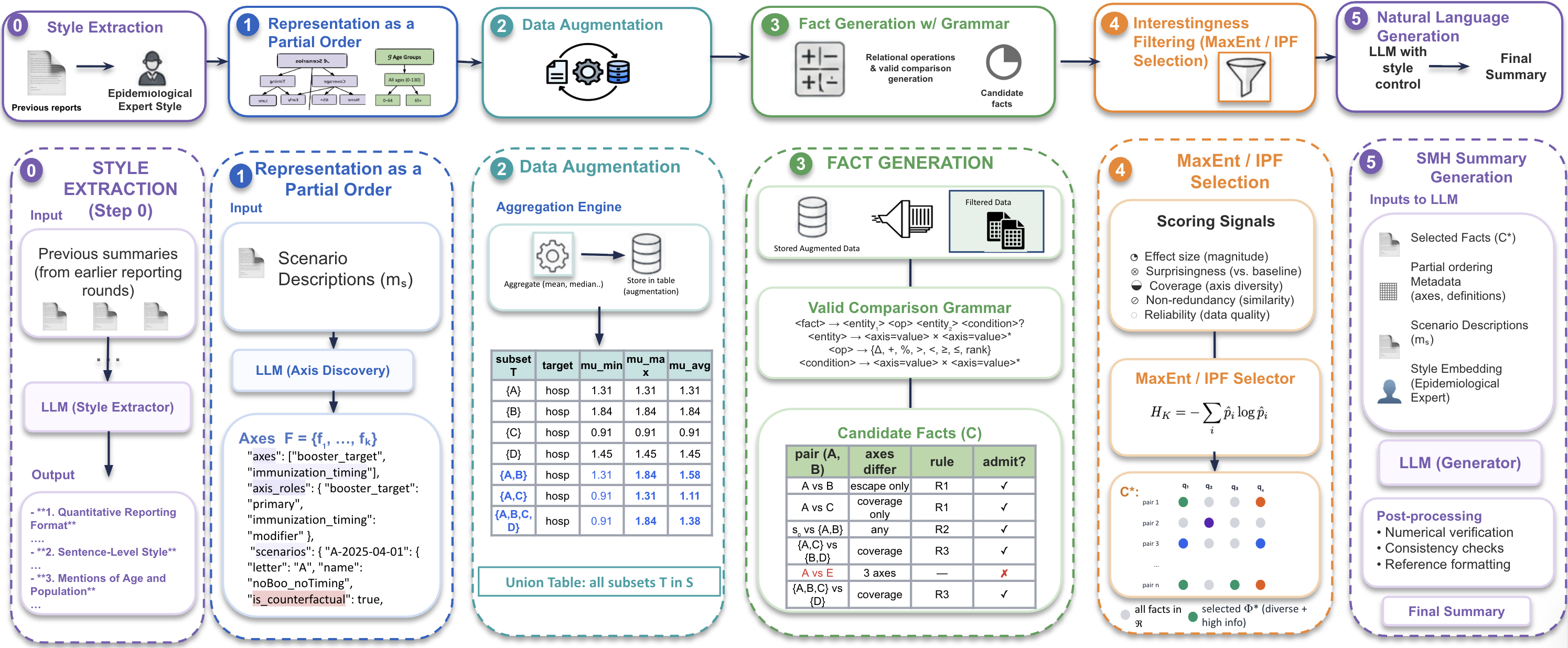}
\caption{Pipeline for scenario-based summary generation. The framework converts projection data into structured scenario axes, builds a multidimensional hypercube, augments subset comparisons, filters valid facts using comparison grammar, MaxEnt and summarizes selected facts with LLM.
}
    \label{fig:workflow}
\end{figure*}

We show that directly prompting even state-of-the-art LLMs to generate summaries from epidemiological data leads to frequent structural inconsistencies and fragile behavior;
this appears to be the case even under few shot learning.
We introduce \tool{}, an agentic framework for generating narratives for public health tasks;
a crucial component in \tool{} is a partial order on some of the attributes, which helps improve the robustness of the generated summaries, as illustrated in Figure~\ref{fig:workflow}.
Our main contributions are: 
\textbf{1.} We formalize \tool{}, a public-health narrative generation problem requiring correctness, coverage, and interestingness; \textbf{2.} We propose \tool{}, an agentic pipeline that uses partial-order-guided schema discovery, dataset augmentation, and comparison grammar to generate grounded facts, and \textbf{3.} We introduce a MaxEnt based interestingness-aware fact selection approach, which is evaluated by both agentic assessment and domain-expert review.

\section{Related Work}

\paragraph{Data-to-text generation.}
Classical data-to-text (D2T) systems map structured records to text through hand-designed pipelines for content selection, sentence planning, and surface realisation~\cite{reiter2000building}. Neural D2T~\cite{lebret2016neural, wiseman2017challenges, parikh2020totto} replaced these stages with end-to-end sequence-to-sequence models trained on record-and-narrative pairs. For example, WikiBio~\cite{lebret2016neural} works on biographical infoboxes, RotoWire~\cite{wiseman2017challenges} on basketball box scores, and ToTTo~\cite{parikh2020totto} on highlighted Wikipedia table cells. Castro Ferreira et al.~\cite{ferreira2019neural} compared end-to-end neural models against pipeline systems and reported that explicit intermediate representations produce more accurate output. All of these methods were built for small flat tables that fit inside the encoder context window rather than the multi-million-cell hypercube that public-health ensemble projections produce.

\noindent
\textbf{Content selection and planning.}
A separate strand of D2T argues that explicit modelling of \emph{what to mention}, and \emph{in what order}, improves faithfulness. Puduppully et al.~\cite{puduppully2019data} introduced a neural pipeline for sports-summary generation that selects records, orders them with a planner, and then realises them as prose. Later work added hierarchical planners~\cite{rebuffel2020hierarchical} and entity-centric plans~\cite{puduppully2019entity}. These planners were trained on small flat record tuples. Adapting them to multi-axis structured inputs requires substantial new engineering as done here.

\noindent
\textbf{LLM prompting over structured inputs.}
Recent work uses LLMs directly as generators, with the structured input flattened or summarised into the prompt. Dibia's LIDA~\cite{dibia2023lida} builds a column-level statistical profile of a dataset such as dtypes, ranges, means, and sample values, and feeds that profile to the model as entire input. Other work serialises tables into prompts as Markdown or CSV fragments and relies on long-context models to reason over the linearised data~\cite{chen2023large,sui2024table}. A common limitation is that the model must decide which contrasts and quantities to surface with no structural constraint on validity, which can produce fabricated or mismatched comparisons.

% \noindent
% \textbf{Chain-of-thought and decomposition prompting.}
% Another line of work tries to improve LLM output through prompt structure rather than input format. Chain-of-thought (CoT) prompting~\cite{wei2022chain} adds an explicit reasoning trace to the prompt. Self-consistency~\cite{wang2023selfconsistency} samples multiple chains and takes a majority vote. Least-to-most prompting~\cite{zhou2023leasttomost} and decomposed prompting~\cite{khot2023decomposed} break a task into sub-questions resolved in order. For numerical reasoning, Program-of-Thoughts~\cite{chen2023large} replaces natural-language reasoning with executable code so that arithmetic is performed deterministically. These methods improve multi-step reasoning. They still rely on the LLM to choose the intermediate computations and to get them right.

\noindent
\textbf{Structured summarization.}
Some recent summarization work induces an intermediate structured representation between input and summary, including structured-representation summarization~\cite{balachandran2021structsum} and graph-aware summarization~\cite{koncelkedziorski2019text}. \tool{} also uses an explicit intermediate structure. It differs in two ways for the public-health setting. First, the intermediate representation is a poset over scenario axes induced from free-text scenario descriptions, rather than a generic graph or knowledge structure. Second, the fact-generation step is governed by a domain-aware comparison grammar that prevents structurally invalid contrasts. We are not aware of any prior summarization system that imposes grammar of this kind, or that has been evaluated on hypercube-scale ensemble projections.

\section{Problem Statement}
We consider the problem of automatically generating natural-language summaries from large structured numerical datasets.
While this is a very general problem, we use public health as an exemplar application.
Let $\mathcal{D}$ denote a dataset organized along $K$ discrete dimensions $\mathcal{F}=\{a_1,\ldots,a_K\}$, where each dimension $a_k$ takes values in a finite set $\mathcal{V}_k$. Examples of such dimensions include intervention scenarios, outcome measures, population subgroups, geographic units, time periods, and uncertainty quantiles. 
For example if $a_k$ denotes age group then $\mathcal{V}_k$ could be the groups $[0-10],  [10-25], \cdots, [65+]$.
Input $\mathcal{D}$ is often represented as a flat table, each record corresponds to a unique combination of dimension values. Thus, $\mathcal{D}$ can be viewed as a $K$-way tensor,
$\quad 
\mathcal{D}: \mathcal{V}_1 \times \cdots \times \mathcal{V}_K \rightarrow \mathbb{R},
$
where each entry $\mathcal{D}[\mathbf{v}]$ stores a numerical statistic associated with one configuration of the dimensions. 

\noindent
\textbf{Problem statement.}
Given $\mathcal{D}$, and optionally a set of textual descriptions $\{m_s\}_{s \in \mathcal{S}}$ describing the scenarios, the goal  is to generate a concise natural-language summary $\mathcal{C}(\mathcal{D})$ that satisfies three criteria:\\
(1) \textbf{Correctness}:
Every quantitative assertion in $\mathcal{C}(\mathcal{D})$ should be grounded in a deterministic computation over $\mathcal{D}$. Formally, for each numerical claim, there should exist an executable computation path $\pi_i$ such that applying $\pi_i$ to $\mathcal{D}$ reproduces the stated value.\\ 
(2) \textbf{Coverage.}
The summary should cover the major dimensions of the dataset. In addition, when reference summaries are available, the generated summary should recover the key information units expressed in the ground-truth summaries associated with $\mathcal{D}$, such as major scenario contrasts, high-burden subgroups, regional heterogeneity, and peak or cumulative trends.\\
(3) \textbf{Interestingness and non-redundancy.}
The objective is to choose a compact subset $\mathcal{C}^{*} \subset \mathcal{C}$, with $|\mathcal{C}^{*}| \ll |\mathcal{C}|$, that is diverse, non-redundant, and prioritizes comparisons with large effect sizes or surprising structure. Redundant claims consume summary budget without improving the reader's understanding of the dataset.

For a given instance, we will output both $\mathcal{C}$ and $\mathcal{C}^*$.
The correctness requirement is critical because LLMs applied directly to large tables often hallucinate statistics, invert rankings, or report incorrect magnitudes.
The coverage requirement ensures that the summary is not limited to scenario-level differences while omitting important variation across subgroups, regions, time, or uncertainty levels.

Jointly satisfying all these is challenging. 
Even a moderately sized setting with 5 scenarios, 3 outcomes, 4 subgroups, 50 regions, 52 weeks, and 23 uncertainty quantiles yields approximately $1.6 \times 10^6$ entries, far exceeding the context budget of current large language models.
Any summarization system must therefore perform selection, aggregation, or compression before generation, and errors at this intermediate stage directly constrain the quality of the final output. Second, the combinatorial structure of $\mathcal{D}$ induces a large space of possible comparisons, making exhaustive enumeration and ranking impractical without a principled selection mechanism. Consequently, naive end-to-end LLM approaches often fail along all three axes: they hallucinate or misstate numerical values, omit important dimensions, and overemphasize redundant or obvious findings.
\section{Method}
\label{sec:method}
We propose a structured pipeline to generate grounded, quantitatively accurate summaries from scenario-based ensemble projections. The pipeline decomposes summary generation into five modular stages: (\emph{i})~latent axis decomposition and dimension discovery, (\emph{ii})~hypercube construction, (\emph{iii})~dataset augmentation via relational operations, (\emph{iv})~fact generation through structured comparison, and (\emph{v})~interestingness filtering via maximum-entropy selection, detailed in Algorithm \ref{ssec:alg}, Appendix. By separating structured numerical reasoning from natural language generation, our framework reduces hallucination, enforces valid comparisons, and preserves quantitative consistency.

\paragraph{Step 0. Style Extraction.}
\label{sec:style}
To ensure the generated summaries maintain stylistic continuity with domain-specific standards, we employ a preliminary style alignment phase (referred to here as Step 0). In this phase, the LLM is provided with summaries from previous reporting rounds to extract a latent representation of the "Epidemiological Expert" tone.

The primary objective of this extraction is to align the linguistic distribution of the key takeaways with the ground truth. This alignment is critical for downstream evaluation; standard lexical metrics such as ROUGE and BLEU often penalize summaries that are factually and numerically accurate but stylistically divergent. By normalizing the output style before fact generation and comparison extraction, we minimize variance attributable to phrasing and improve the sensitivity of our automated evaluation metrics.

\paragraph{Step 1. Representation as a Partial Order.}

We represent the dataset as a relation $\mathcal{A}(a_i: \mathcal{V}_i, i=1,\ldots, K)$, where $a_i$ represents an attribute and $\mathcal{V}_i$ represents its domain;
$\mathbf{t}\in \mathcal{V}_1\times \ldots \mathcal{V}_K$ correspond to possible type values. Rather than assuming a fixed schema, we induce the underlying factor structure directly from free-text scenario descriptions using an LLM $\mathcal{M}$.

In the public health dataset we consider, attributes include interventions (e.g., vaccination policies and compliance), demographics (e.g., age groups), regions (e.g., states), and specific metrics (e.g., infection and hospitalization rates).
We sometimes refer to type values $\mathbf{t}$ as scenarios and attributes $a_i$'s as scenario axes.
We use $\mathbf{t}_i$ to refer to the value of the $i$th scenario axis.
\begin{wrapfigure}{r}{0.42\linewidth}
    \centering
    \vspace{-0.3cm}
    \includegraphics[width=\linewidth]{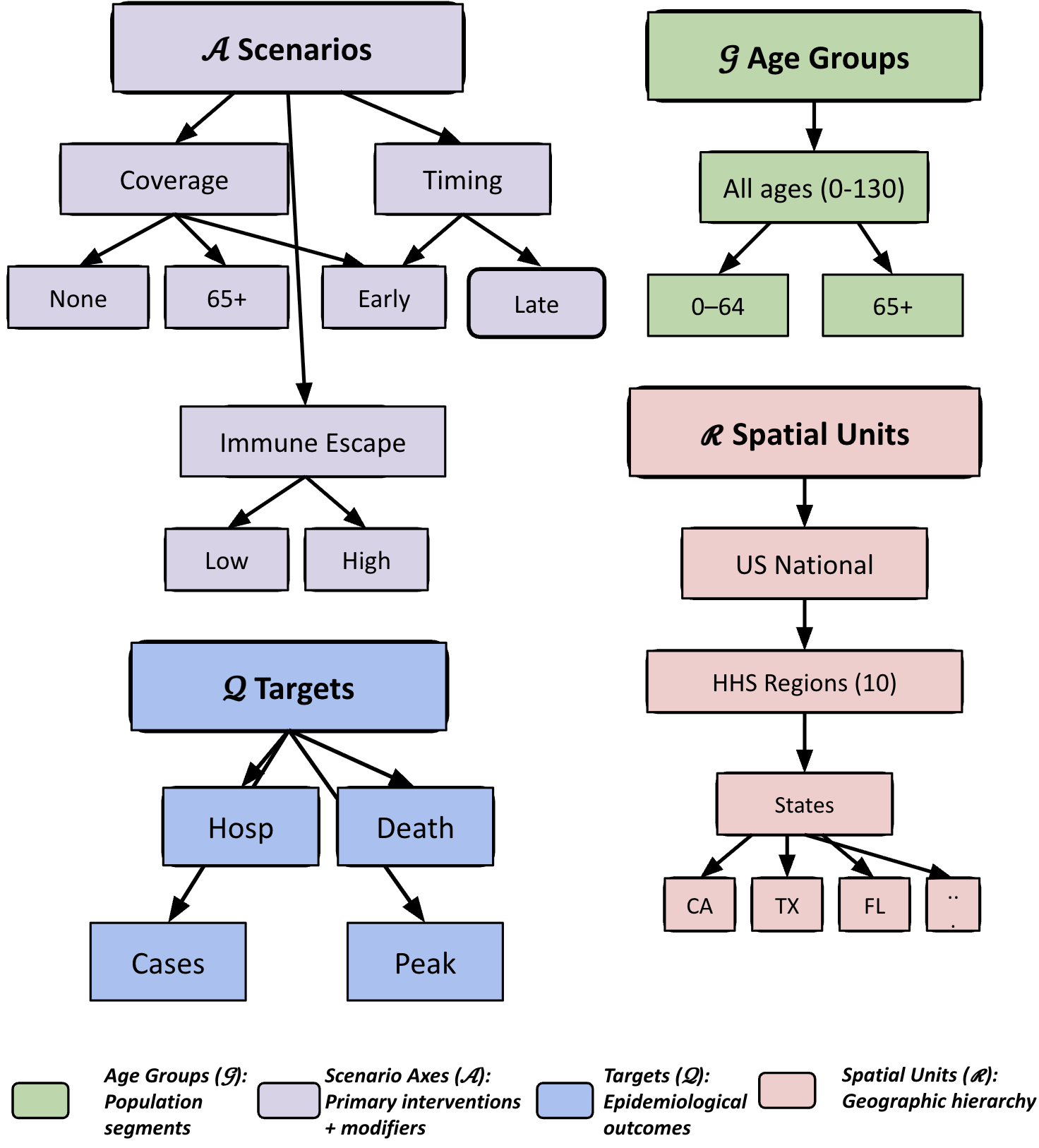}
    \caption{Partial order over attributes.}
    \label{fig:partial-order}
    \vspace{-0.5cm}
\end{wrapfigure}
There can be a partial order (or poset) $\leq_i$ on each attribute $a_i$.
For instance, the intervention attribute for public health can consist of specific vaccination policies (e.g., vaccination rates for different demographics) and timing (when the policy is implemented).
The natural hierarchy among these intervention types forms a partial order (Figure \ref{fig:partial-order}). These might additionally be classified as ``primary'' (specifying  which policy or strategy is applied) and ``modifier'' (specifying its implementation, e.g., timing or intensity).
Similarly, the hierarchy associated with demographics forms a poset.
The posets for different attributes can be combined to form a poset on the relation $\mathcal{A}$.
% \anil{describe how this is being done by agents} \ritu{done}

In our pipeline, this structure is constructed and exploited by two agents. A \emph{decomposition agent} $\mathcal{M_D}$ maps each free-text scenario description to a structured tuple $\mathbf{t}\in\mathcal{V}_1\times\cdots\times\mathcal{V}_K$, distinguishing primary from modifier axes.
A \emph{comparison-selection agent} then traverses the induced product poset to identify three classes of structurally valid comparisons — defined formally in Step~3(\S \ref{sec:facts}).

\noindent
\textbf{Robust schema inference.} 
Data formats are not fully consistent and uniform.
We use an  agentic component to learn a poset representation for the dataset. 
The model is provided with a statistical profile of the dataset–field names, data types, cardinalities, and sample values–to produce a canonical internal representation $\sigma$.
If schema loading fails or raises a validation error $\epsilon^{(t)}$,
the representation is corrected iteratively:
$\small
    \sigma^{(t+1)} \;\leftarrow\; \mathcal{M}\!\left(\sigma^{(t)},\,
    \epsilon^{(t)}\right),
    \label{eq:schema-loop}
$\\
up to a maximum of $T$ attempts. This allows automatic diagnosis and correction of type mismatches or incorrect column identifications without manual intervention.
Let $\mathcal{D}$ denote the learned dataset.
For each datapoint $\mathbf{x}\in\mathcal{D}$, let $\mathbf{t}(\mathbf{x})$ denote the corresponding type descriptor.

\paragraph{Step 2. Dataset Augmentation.}
\label{sec:augmentation}

Starting from $\mathcal{D}$, we construct $\mathcal{D}_{\mathrm{aug}}$ using a sequence of operations, and augment it via deterministic relational operations.
Examples of operations are \emph{unions}.
A union of a subset $S\subset \mathcal{D}_{aug}$ can be done if there exists an attribute $i$ such that:
(1) for all $j\neq i$, $\mathbf{t}(\mathbf{x})_j, \mathbf{x}\in S$ are the same, and
(2) there exists an item $y\in \mathcal{V}_i$ which is a parent of $\mathbf{t}(\mathbf{x})_i, \mathbf{x}\in S$.
Data points corresponding to such a subset $S$ can be combined to form an additional data point $\mathbf{x}(S)$ with a value that depends on how they can be aggregated; we set the associated type $\mathbf{t}(\mathbf{x})_i=y$.
For instance, the value in $\mathbf{x}(S)$ might be a sum of the values of the individual items, or can correspond to the associated range (between the minimum to the maximum).
An agent is implemented to apply the grammar rules and  combine such subsets of items (with chain of thought reasoning to ensure correctness), which are added repeatedly to the dataset $\mathcal{D}_{aug}$.

\paragraph{Step 3. Fact Generation.}
\label{sec:facts}
We now specify a grammar for generating facts
from $\mathcal{D}_{\mathrm{aug}}$.
This is done in two ways.
First, each $\mathbf{x}\in \mathcal{D}_{aug}$ can be translated to a fact.
Second, we generate a set of facts with quantitative claims about relationships between scenarios, using a \emph{Comparison Grammar} that restricts which pairs are valid, preventing spurious ``apples-to-oranges'' analyses. 
It helps the LLM by providing a structured guideline for generating the summary, instead of relying entirely on the LLM’s random or unconstrained generation.

Examples of rules satisfied by valid comparisons in the grammar are:
(\textbf{R1})~two scenarios differ on \emph{exactly one} axis while agreeing on all others; (\textbf{R2})~the baseline scenario is compared against a pool of active-condition scenarios that all share at least one common secondary axis value; or (\textbf{R3})~two such pools are compared only when their members are matched on every axis except the one of interest. 
Each comparison generated by the grammar is associated with a quantity such as the percentage change in mean summary statistic between the two sides. 
We use $\mathcal{C}$ to denote the set of all outputs generated by the comparison grammar; these are translated to facts in the narrative.
This ensures that every fact is numerically grounded and tied to a specific attribute.

\color{black}

% \subsection{Step 4. Interestingness via Maximum Entropy Selection}
\paragraph{Step 4. Generate interesting facts.}
\label{sec:maxent}

The set $\mathcal{C}$ of facts may contain redundant or low-information
comparisons. 
There are many ways in which a statement can become interesting, e.g., changes in temporal and spatial patterns, or relative to reports in the literature.
We formalize interestingness selection using a maximum-entropy (MaxEnt) approach \cite{jaynes1957information} and construct a compact, high-salience subset
$\mathcal{C}^* \subseteq \mathcal{C}$.
% \anil{some details}
Each comparison $c = (A, B) \in \mathcal{C}$ is associated with a \emph{feature profile} $\phi(c) \in \mathcal{F}$ that records which combination of scenario axes, target quantity, subgroup, spatial unit, and any other indexed dimension it addresses.
These are computed using Iterative Proportional Fitting (IPF), which alternately normalizes the distribution to satisfy each dimension's constraint in turn and converges to the maximum-entropy solution. 
This discourages redundancy (e.g., multiple comparisons addressing the same target or subgroup) while rewarding facts that illuminate underrepresented cells of $\mathcal{C}$.

For state-level interestingness, we instantiate this selection problem over a state-by-field observation matrix, where each cell represents a normalized projection statistic for a state and epidemiological field.
Let $\mathbf{M} \in \mathbb{R}^{N \times P}$ be an observation matrix where $M_{ij}$ is the point-estimate projection for state $i$ on epidemiological field $j$ ($N$ U.S. states, $P$ fields from SMH Round). Fields span heterogeneous units, so we column-normalise: $\tilde{M}_{ij} = \frac{M_{ij} - \mu_j}{\sigma_j}$\\
\textit{The goal is to identify cells $(i,j)$ whose values are surprising and not explained by state-level effects alone.} We evaluate the model for the expected value $F_{ij}$, against which the residual $R_{ij} = \tilde{M}_{ij} - F_{ij}$ defines surprise.

\paragraph{MaxEnt/IPF.}
We fit a maximum-entropy matrix $\mathbf{F}$ using Iterative Proportional Fitting (IPF), starting from a uniform prior and constraining it to match the observed marginal totals for all regions, field categories, and outcome metrics. To penalise redundant findings, we measure the entropy of the state distribution in the top-$K$ list.
A higher $H_K$ indicates findings spread across more states. $H_K = -\sum_i \hat{p}_i \log \hat{p}_i$

\paragraph{Final Summary Generation.}
Once the top surprising cells are identified, we generate a written summary using a large language model (LLM). Rather than prompting the LLM directly with raw numbers, we use a three-step chain-of-thought (CoT) pipeline to ensure structured reasoning.\\
\textbf{Step A: Observe} The LLM is given a structured findings block containing the top-$K$ surprising $(i, j)$ cells, their raw values, residuals, and regional aggregates. It is asked to describe what is driving each state's surprise score and identify any regional or category-level patterns.\\ 
\textbf{Step B: Reflect} The LLM is given its Step A output and asked to select the 2–3 most policy-relevant findings. For each, it produces a one-sentence finding, an explanation of why the deviation is epidemiologically meaningful.\\
\textbf{Step C: Synthesize} Using the observations from Steps A and B, the LLM writes a concise national executive summary in plain language, suitable for a public health audience.

The three-step structure serves two purposes.
First, it separates pattern detection (Step A) from interpretation (Step B) from communication (Step C), reducing the chance that the LLM fabricates explanations for statistical artifacts. Second, each step's output is logged alongside its prompt, making the reasoning chain understandable. 
\section{Experimental Setup}

\textbf{Data.}
We evaluate our framework on the \textbf{Scenario Modeling Hub (SMH)} dataset (referred to as GT), a multi-round ensemble modeling initiative for infectious disease projections. 
We specifically focus on COVID-19 projections, where each modeling round poses a structured set of scenarios defined along multiple axes: e.g., booster coverage (none, 65+, all ages), immune escape (low, high), and timing (classic, early), resulting in 4–6 distinct scenarios per round. Each scenario is projected by multiple independent modeling teams, which are then ensembled to produce quantile outputs over the projection horizon.

\noindent
\textbf{Dataset characteristics.}
\emph{Scenario descriptions}: Free-text markdown files (150–400 words per scenario) on modeling assumptions, booster coverage and immune escape parameters (
for the most recent round), and related epidemiological context.\\ \emph{Ensemble projections}: Probabilistic (quantile-based) outputs per scenario across target quantities (hospitalizations, deaths), stratified by age group ($|\mathcal{G}| = 2$: 0–64, 65+) and jurisdiction ($|\mathcal{R}| = 51$ U.S. states + national).
\emph{Reference summaries.} Expert-curated national executive summaries (400–800 words) as well as templatized state-level summaries, serving as ground truth for evaluation.

\noindent
\textbf{Implementation Details.}
All LLM calls use \textbf{Gemma 4 26B} (gemma-4-26b-a4b-it) \cite{google2026gemma4} via the Google GenAI API free tier, with temperature 0.0 for reasoning and 0.2–0.4 for generation; the state-level MaxEnt summary uses \textbf{Gemma 4 31B}, no fine-tuning is performed. For interestingness analysis, we use ipfn package to implement IPF; M3 metrics use scikit-learn, and embedding similarity uses sentence-transformers (all-MiniLM-L6-v2) \cite{reimers2019sentencebert}. All prompts and responses are logged to JSONL for auditability. The full pipeline runs in 3–5 minutes per round, dominated by API latency.

\section{Results}

We report our results across two distinct dimensions. While NLP metrics (BLEU, ROUGE) measure adherence to the existing GT, they may paradoxically penalize our framework for identifying "MaxEnt" (Maximum Entropy) insights—critical epidemiological observations that are statistically present in the data but were not reported in the GT.

Because the GT is not an exhaustive gold standard, a summary that captures higher-utility insights may receive lower lexical overlap scores. Consequently, we separate our findings into \textit{Lexical Alignment (closeness to GT)} and Expert Utility (human ranking of "interestingness"). We demonstrate that our framework consistently surfaces important components that experts prefer, even when those components are absent from the manual reports.

% \begin{table}[t]
% \centering
% \caption{Comparison of summary evaluation metrics for two generated summaries.}
% \label{tab:summary_eval}
% \begin{tabular}{lccc}
% \hline
% \textbf{Metric} & \textbf{GPT 5.5} & \textbf{CLAUDE SONNET 4.6} & \textbf{Ours} \\
% \hline
% BLEU-4 & 0.0461 & 0.0277 &\\
% SacreBLEU & 4.5734 & 3.0118 & \\
% ROUGE-1 F1 & 0.4525 & 0.3883 &\\
% % ROUGE-2 F1 & 0.1173 & 0.0985 \\
% % ROUGE-L F1 & 0.1832 & 0.1622 \\
% ROUGE-Lsum F1 & 0.4216 & 0.3607 &\\
% Sentence Embedding Cosine Similarity & 0.8456 & 0.8195 & \\
% % BERTScore F1 & -0.0059 & -0.0248 \\
% \hline
% \end{tabular}
% \end{table}
% \begin{table}[t]

\subsection{Evaluation Metrics}

We evaluate generated summaries using three complementary metrics: factual fidelity,
coverage of ground-truth findings, and stylistic consistency with the reference report.
\begin{wraptable}{r}{0.78\textwidth}
\centering
% \vspace{-1cm}
\caption{Comparison of M1, M2, and M3 scores across SMH rounds. For M1, ``100\%/30'' denotes 100\% factual fidelity over 30 extracted numerical claims.}
\label{tab:model_scores_rounds_compact}
\resizebox{\linewidth}{!}{
\begin{tabular}{lccc ccc ccc}
\toprule
\textbf{Model}
& \multicolumn{3}{c}{\textbf{R19}}
& \multicolumn{3}{c}{\textbf{R18}}
& \multicolumn{3}{c}{\textbf{R17}} \\
\cmidrule(lr){2-4}
\cmidrule(lr){5-7}
\cmidrule(lr){8-10}
& \textbf{M1} & \textbf{M2} & \textbf{M3}
& \textbf{M1} & \textbf{M2} & \textbf{M3}
& \textbf{M1} & \textbf{M2} & \textbf{M3} \\
\midrule
\rowcolor{decoderblue}
\multicolumn{10}{l}{\textit{Decoder-based / LLM baselines}} \\

LIDA
& 38\%/8 & 0.17 & 0.50
& 20\%/10 & 0.00 & 0.42
& 25\%/8 & 0.00 & 0.57 \\

Single-shot CoT
& 0\%/4 & 0.00 & 0.43
& 12\%/8 & 0.00 & 0.48
& 83\%/6 & 0.50 & 0.38 \\

Multi-stage CoT
& 60\%/5 & 0.42 & 0.45
& 0\%/5 & 0.00 & 0.46
& 40\%/5 & 0.00 & 0.50 \\

Plan-then-Generate
& 0\%/6 & 0.33 & 0.48
& 20\%/5 & 0.00 & 0.46
& 20\%/5 & 0.00 & 0.60 \\

GPT 5.5
& \textbf{100\%}/13 & 0.35 & \textbf{0.79}
& 88\%/28 & 0.40 & \textbf{0.86}
& \textbf{100\%}/15 & 0.28 & \textbf{0.85} \\

Claude Opus 4.6
& 0.9\% & 0.38 & 0.09
& 60\%/32 & 0.38 & 0.80
& 75\%/23 & 0.25 & 0.82 \\

\midrule
\rowcolor{encoderorange}
\multicolumn{10}{l}{\textit{Encoder--decoder baselines}} \\

T5-Large
& 33\%/3 & 0.33 & 0.34
& 67\%/3 & 0.00 & 0.34
& 100\%/1 & 0.00 & 0.48 \\

BART
& 0\%/1 & 0.00 & 0.34
& 0\% & 0.00 & 0.06
& 0\%/1 & 0.00 & 0.38 \\

\midrule
\rowcolor{oursgreen}
\multicolumn{10}{l}{\textit{Ours}} \\

\tool{}
& \textbf{100\%}/30 & \textbf{0.78} & 0.69
& \textbf{89\%}/34 & \textbf{0.44} & 0.81
& \textbf{100\%}/26 & \textbf{0.62} & 0.79 \\

\bottomrule
\end{tabular}
}
\vspace{-1em}
\end{wraptable}
\noindent
\textbf{Factual accuracy (M1).}
M1 measures whether the generated summary makes numerically correct claims.
We use a feedback LLM to extract quantitative assertions from the generated summary and verify each claim against the source projection tables.
We also check whether a number is directly reported or derived, since values such as percentage increases or decreases may not appear explicitly in the table.
The final score is the fraction of verified numerical claims: $ \mathrm{M1} =\frac{ N_{\mathrm{verified}}}{ N_{\mathrm{claims}}}.$

\noindent
\emph{Caveat.} Since M1 is based on LLM-extracted numerical claims and LLM-based verification, the score can be stochastic and imperfect.
In some cases, a correctly derived value may be marked as incorrect if the verifier fails to trace the arithmetic or match the derived number to the source tables (Appendix \ref{fig:M1_issue}).
Therefore, M1 should be interpreted as an approximate factual-fidelity measure rather than an exact ground-truth score.

Another limitation is that M1 does not directly capture the amount of quantitative evidence included in the summary (Table \ref{tab:model_scores_rounds_compact}).
For example, a summary that contains only one numerical claim can receive a perfect M1 score if that claim is correct, even though it may be less informative than a number-rich summary with broader quantitative coverage.

\noindent
\textbf{Coverage (M2).}
M2 measures how well the generated summary captures the main scenario-based findings from the ground-truth (GT) Key Takeaways.
The evaluator first extracts distinct GT information units, such as intervention effects, marginal gains, absolute burden under named scenarios, and age-group burden patterns.
Each GT point is then marked as covered, partially covered, or missing in the generated summary. We then compute:
$$\small
\mathrm{M2}
=
\frac{
    N_{\mathrm{covered}} + \frac{1}{2}N_{\mathrm{partial}}
}{
    N_{\mathrm{GT}}
}.
\label{eq:m2_coverage}$$
\noindent
\textbf{Style consistency (M3).}
M3 measures whether the generated summary follows the writing style of the GT Key Takeaways.
We compare the generated and GT summaries using cosine similarity over their text representations.
In implementation, this is computed primarily using embedding similarity.
% \vspace{-1em}
$$\small
\mathrm{M3}
=
\cos\!\left(
    \mathbf{e}_{\mathrm{gen}},
    \mathbf{e}_{\mathrm{GT}}
\right)
=
\frac{
    \mathbf{e}_{\mathrm{gen}}^{\top}\mathbf{e}_{\mathrm{GT}}
}{
    \left\lVert \mathbf{e}_{\mathrm{gen}} \right\rVert_2
    \left\lVert \mathbf{e}_{\mathrm{GT}} \right\rVert_2
}.
\label{eq:m3_style}
$$

Table~\ref{tab:model_scores_rounds_compact} shows that \tool{} achieves strong M1 scores across rounds while making a larger number of numerical claims, which is desirable for quantitative public-health summaries. \tool{} also achieves the highest M2 coverage, suggesting that the partial-order representation helps recover more valid scenario-axis comparisons than unconstrained baselines. For M3, \tool{} remains comparable while providing stronger quantitative grounding and coverage.

\subsection{MaxEnt-based Interestingness Analysis}
\label{sec:maxent_results}

To identify state-level patterns that are not apparent from aggregate national summaries alone, we apply a maximum-entropy interestingness analysis to the state-level projection outputs. The MaxEnt baseline is fitted using iterative proportional fitting (IPF), simultaneously matching HHS regional marginals (row axis) and field-category/metric marginals (column axis).
All absolute burden and averted-count fields are expressed per 100{,}000 population using 2022 ACS five-year estimates before the matrix is constructed, so that state size does not mechanically inflate residual
scores.
Large absolute residuals indicate state--metric cells that are surprising relative to what would be expected from the marginal structure alone. We assign states to HHS regions, which group U.S. states and territories into ten regions~\cite{hhs_regions}.

\begin{wrapfigure}[17]{r}{3.8in}
\centering
\vspace{-2em}
\includegraphics[width=\linewidth]{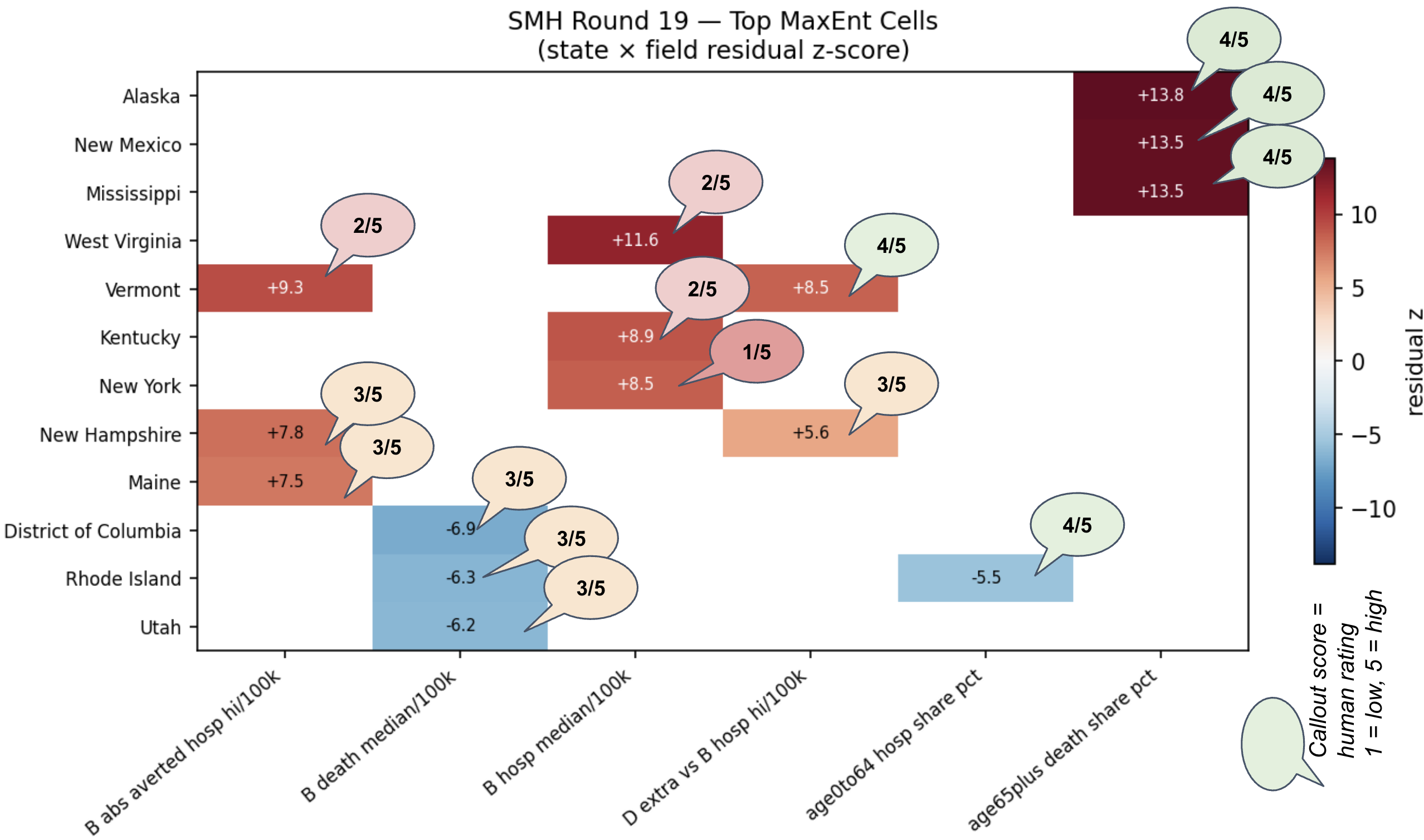}
\caption{\footnotesize MaxEnt-ranked residual cells for R19 with human interestingness ratings.}
\label{fig:maxent-r19}
\end{wrapfigure}

\emph{Evaluation and justification.} Figure~\ref{fig:maxent-r19} shows the top MaxEnt cells with human interestingness ratings (Details in Appendix Fig, \ref{fig:maxent1},\ref{fig:maxent2},\ref{fig:maxent3},). High positive residuals highlight above-expected values, while negative residuals indicate below-expected values. Several high-ranked cells involve age-specific mortality shares: Alaska, New Mexico, and Mississippi have higher-than-expected shares of projected deaths among adults aged 65+. Other high-ranked findings capture elevated hospitalization burden in West Virginia, Kentucky, and New York, as well as stronger-than-expected intervention effects on averted hospitalizations in Vermont, New Hampshire, and Maine. DC, Rhode Island, and Utah show lower-than-expected death burden.

Expert ratings suggest that MaxEnt can be useful for state-level signals, but not all statistically surprising cells are equally narrative-worthy. Age-share findings generally receive higher ratings, while some hospitalization-burden findings are rated lower because they may require additional context or may be less policy-informative. Overall, the MaxEnt analysis provides a complementary selection mechanism for identifying state-level deviations, while human grading helps distinguish statistically surprising cells from genuinely useful narrative content.

% \subsubsection{Expert Utility}
% \paragraph{Human evaluation (M4)}
% - Domain experts: 
% - Rank on: interestingness

\subsection{Baselines}
\label{sec:baselines}

We compare \tool{} against five families spanning the data-to-text design space, along with a schema-only information-deprived floor. All baselines are evaluated under the same protocol on rounds R17, R18, and R19. Table~\ref{tab:baselines} summarizes the baseline categories, implementation details, and main observed limitations.

\begin{table*}[t]
\centering
\caption{Summary of baseline families, implementation choices, and observed limitations.}
\small
\resizebox{\linewidth}{!}{%
\begin{tabular}{|p{0.10\linewidth}| p{0.55\linewidth}| p{0.55\linewidth}|}
\toprule
\textbf{Baseline family} & \textbf{Implementation} & \textbf{Observed limitation} \\
\midrule
Schema-only LLM \cite{dibia2023lida}
&
The LLM receives only a column-level statistical profile of $\mathcal{D}$, including dtypes, ranges, means, and sample values, with no row-level access.
&
\textbf{1.} Provides less information. \textbf{2.} In practice, LIDA can hallucinate magnitudes from sampled cell values, e.g., reporting the global \texttt{value.max()} as if it were a valid scenario projection. \\

\midrule

Direct LLM prompting \cite{achiam2023gpt, openai2026gpt55,anthropic2026claudeopus46}
&
The hypercube $\mathcal{D}$ is serialized as a CSV using a stratified random subsample over scenario $\times$ target $\times$ age $\times$ location, restricted to the $\{0.025, 0.5, 0.975\}$ quantiles, and supplied to a long-context LLM with scenario descriptions $\{m_s\}$.
&
\textbf{1.} GPT-5.5 is factually conservative but misses several headline findings, such as the two-peak trajectory and key R19 magnitudes. \newline
\textbf{2.} Claude Opus 4.6 is more brittle, in R19, misidentifies the report as influenza study and shifts projection start by a month. \\

\midrule

Chain-of-thought prompting
\cite{wei2022chain,wang2023selfconsistency,zhou2023leasttomost,khot2023decomposed}
&
We evaluate two CoT regimes: single-shot CoT, using a single prompt that asks the model to identify contrasts, compute percentage changes, and write the narrative; and multi-stage CoT, which follows a question generation $\rightarrow$ per-question reasoning $\rightarrow$ synthesis pipeline.
&
\textbf{1.} CoT improves some arithmetic reasoning, but without an explicit scenario representation or comparison grammar, it can still \emph{miss valid scenario-axis comparisons} or reason over invalid contrasts. \\

\midrule

Content selection and planning
\cite{puduppully2019data}
&
The original Plan-then-Generate repository targets RotoWire boxscores and does not directly support hypercube-scale inputs. We therefore re-implement the plan-then-realize architecture as a two-call LLM cascade using the same backbone for fair comparison.
&
\textbf{1.} The realization stage sometimes drops scenario identifiers from planned facts, e.g., stating ``the model projects 1{,}707 deaths'' without specifying the scenario, which prevents the verifier from anchoring claims to valid scenario contrasts. \\

\midrule

Fine-tuned seq2seq models
\cite{ferreira2019neural,raffel2020exploring,lewis2020bart}
&
Only $n=3$ SMH rounds are publicly available with paired Key Takeaways, which is insufficient for fine-tuning. We therefore evaluate T5-Large and BART zero-shot on the textual data context, which is the strongest representation they can consume within their 512--1024 token input budgets.
&
\textbf{1.} T5-Large shows an extractive-copying artifact, it copies one data-context line before collapsing into incoherent text. \newline
\textbf{2.} BART collapses almost entirely to broken date/scenario tokens, indicating that pretrained summarization models cannot reliably ingest this representation zero-shot. \\

\bottomrule
\end{tabular}%
}
\vspace{-0.4cm}
\label{tab:baselines}
\end{table*}

Overall, decoder-based LLM baselines often produce fluent summaries and can achieve high surface-level alignment scores, but they are less reliable in factual fidelity and coverage because numerical reasoning and comparison selection are not explicitly constrained. CoT prompting improves some arithmetic reasoning, but without a structured scenario representation, it can still miss valid scenario-axis comparisons or reason over invalid contrasts. Encoder--decoder baselines such as T5-Large and BART tend to produce shorter and more generic summaries with very few numerical claims; this can sometimes inflate $M_1$ when the claim count is small, but generally leads to poor coverage of scenario-based ground-truth findings. In contrast, \tool{} maintains a larger number of grounded numerical claims while achieving the strongest coverage across rounds.

\subsection{Ablation Study}
\label{sec:ablation}

To evaluate the contribution of each component in \tool{}, we conduct an ablation study by removing one module at a time while keeping the rest of the pipeline unchanged. We focus on three core components: \textbf{1.} chain-of-thought reasoning, \textbf{2.} partial order, and \textbf{3.} preprocessing/schema agent. The goal is to assess whether each module contributes to factual fidelity, coverage, and alignment with expert-written summaries.

\begin{table}[t]
\caption{Ablation study of \tool{} components}
\label{tab:ablation_study}
\resizebox{\textwidth}{!}{
\begin{tabular}{|l|rcc|l|}
\toprule
\textbf{Model Variant} & \textbf{M1} & \textbf{M2} & \textbf{M3} & \textbf{Observed Failure Mode} \\
\midrule
\tool{}              & \textbf{100\%} of 30 claims & \textbf{0.78} & 0.69 
& Full structured pipeline \\
$\tool{}_{w/o\ cot}$ & 70\% of 46 claims & 0.52 & 0.68
& Weaker derivation tracing; more numerical inconsistencies \\
\tool{} w/o Partial Order 
& \textbf{100\%} of 27 claims & 0.28 & \textbf{0.73} 
& Invalid or missing scenario-axis comparisons \\

\bottomrule
\end{tabular}
}
\vspace{-1em}
\end{table}

The ablation results in Table \ref{tab:ablation_study} show that CoT mainly improves factual fidelity by helping the model trace and verify derived numerical claims, increasing M1 from 70\% to 100\%. In contrast, the partial-order representation is most important for coverage: without it, the model still produces numerically correct claims, but misses many valid scenario-axis comparisons, causing M2 to drop from 0.78 to 0.28.
\emph{\tool{} w/o Preprocessing Agent} cannot be applied robustly across different SMH rounds, as it leads to more schema and type-mismatch errors. This component is necessary in our pipeline because each round may have different ensemble data formats, scenario definitions, and intervention policies.
  
% \subsection{Ablation Study}

% To evaluate the contribution of each module, we performed an ablation analysis (summarized in Figure \ref{fig:baseline}). Key findings include:
% \begin{itemize}
% \item \textbf{w/o CoT:} Led to a significant increase in numerical hallucinations and logical fallacies.
% \item \textbf{w/o Structured Tuples:} Resulted in invalid or missing scenario comparisons across axes.
% \item \textbf{w/o Feedback Loop:} Compromised the internal consistency of the summary.
% \item \textbf{w/o Preprocessing Agent:} Increased type-mismatch errors during schema loading.
% \end{itemize}
\section{Conclusion}

We introduced \tool{}, an agentic framework for generating grounded and comprehensive narratives from epidemiological scenario-modeling outputs. Rather than prompting an LLM directly over large projection tables, our approach separates structured numerical reasoning from natural-language generation. The framework first decomposes scenario descriptions into structured axes, builds an augmented scenario space, generates valid comparisons through a comparison grammar, and then selects salient findings using interestingness-based filtering, including MaxEnt/IPF.
Across SMH rounds, \tool{} improves coverage of key scenario-based findings while maintaining strong factual fidelity of numerical claims. The results suggest that explicit structure is important for this task. The MaxEnt analysis further shows that the framework can surface state-level and age-specific patterns that may not appear in aggregate national summaries, providing an additional mechanism for identifying potentially policy-relevant findings.\\
\textbf{Limitations and Future Work.} Some statistically surprising MaxEnt signals may require human review to distinguish meaningful patterns from noise; however, this human-in-the-loop step can also strengthen the reliability and usefulness of the generated public-health narratives. Our evaluation is currently limited to scenario-modeling rounds with factorial designs, and the pipeline uses Gemma 4 models accessed through a free-tier API. Performance may therefore vary across tasks, scenario structures, or model backbones. Validating \tool{} across more diverse public-health settings and LLMs remains an important direction for future work.

\bibliographystyle{plain}
\bibliography{refs}
\clearpage
\appendix
\section{Additional Details on Partial Ordering}
\label{app:partial-order}

The partial order is the central structural device in \tool{}. It serves three roles: (i)~guiding the LLM decomposition agent toward a consistent axis representation, (ii)~defining which scenario subsets can be legitimately merged during augmentation, and (iii)~constraining the comparison grammar so that every emitted fact corresponds to a valid, interpretable contrast.

\paragraph{Axis extraction (agentic).}
Each round contains free-text scenario descriptions that implicitly encode the experimental design: which factors vary, which are held fixed, and which scenario serves as the no-intervention baseline. Algorithm~\ref{alg:decomp} recovers this structure by prompting $\mathcal{M}$ to map each scenario to a tuple of axis values and to classify each axis as \texttt{primary} (the intervention itself, e.g.\ vaccination target population) or \texttt{modifier} (a parameter of the intervention, e.g.\ timing). The distinction induces a two-level partial order on the intervention attribute: primary values are incomparable to each other (they represent qualitatively different strategies), while modifier values refine a given primary choice and are therefore ordered beneath it. This hierarchy is what prevents the comparison grammar from pooling across primary axes in grouped-counterfactual comparisons (Rule~R2).

\begin{figure}[t]
    \centering
    \includegraphics[width=\linewidth]{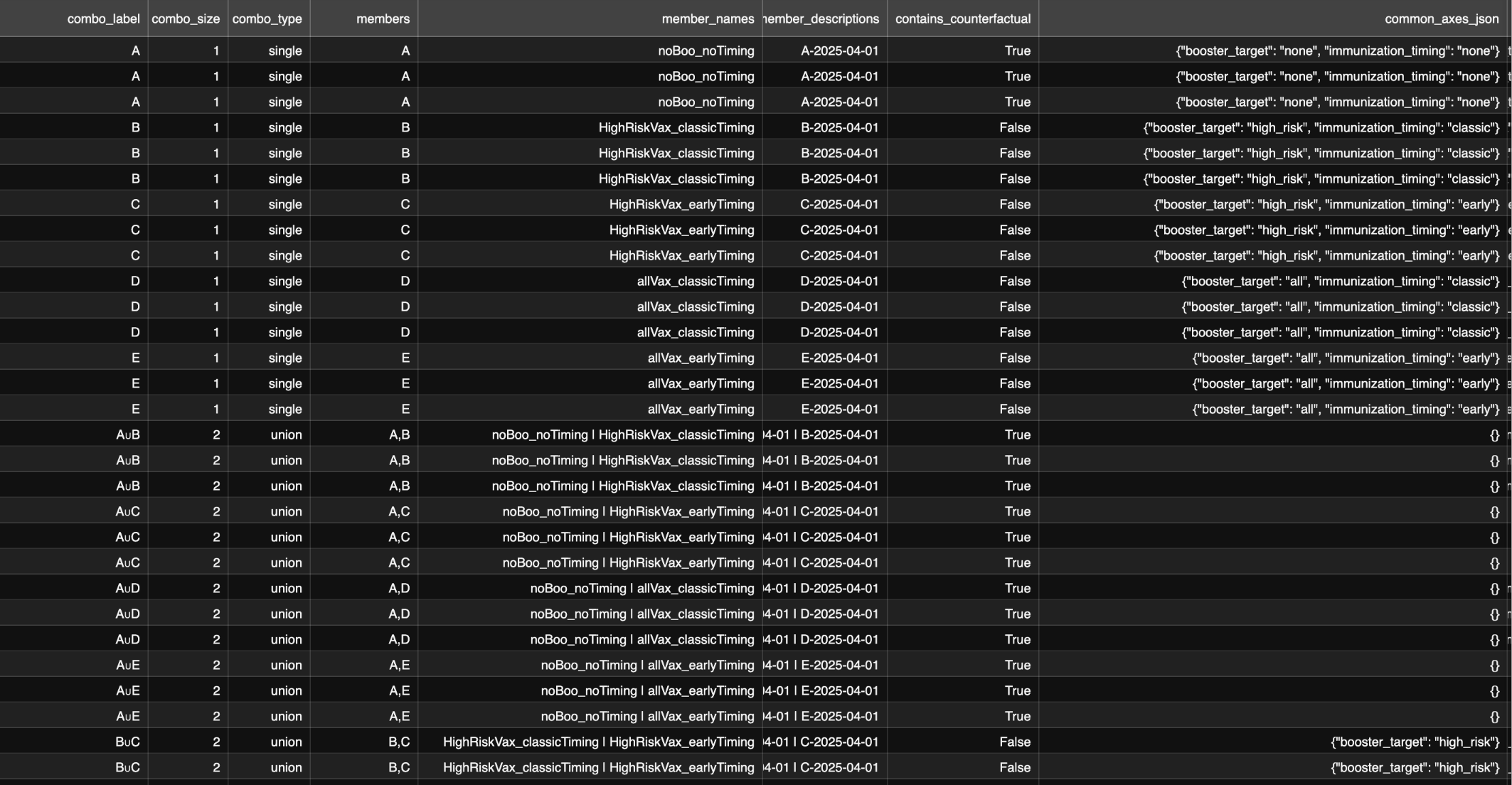}
    \caption{Excerpt from the augmented dataset $\mathcal{D}_{\mathrm{aug}}$ for Round~19. Rows with \texttt{combo\_size}$\,{=}\,1$ are individual scenarios; larger subsets aggregate member-level median projections. The \texttt{common\_\_*} columns show axis values shared by all members of the subset (intersection), while \texttt{union\_\_*} columns list all values present (union). Numeric columns report the mean of member medians for each target, which are used by the comparison grammar to compute percentage changes.}
    \label{fig:augmented}
\end{figure}

\paragraph{Dataset augmentation.}
Given the per-scenario axis tuples, the augmentation step (Algorithm~\ref{alg:augment}) constructs $\mathcal{D}_{\mathrm{aug}}$ by taking all $2^{|\mathcal{S}|}{-}1$ non-empty subsets of the scenario set. A union of subset $C$ is valid when there exists an attribute $i$ such that all members of $C$ agree on every other attribute $j \neq i$ and the value on attribute $i$ has a common parent in the partial order. In practice, the union table records---for each subset---which axis values are shared (common axes, the intersection $\bigcap_{s \in C} \mathbf{t}_s$) and which vary (union axes, $\bigcup_{s \in C} \mathbf{t}_s$). Per-target statistics (min, max, mean of member medians) are attached for every (age group, target) pair. Figure~\ref{fig:augmented} shows an excerpt: singleton rows correspond to individual scenarios; multi-member rows aggregate statistics across the subset, enabling downstream comparisons at different levels of granularity.

\begin{figure}[t]
    \centering
    \includegraphics[width=\linewidth]{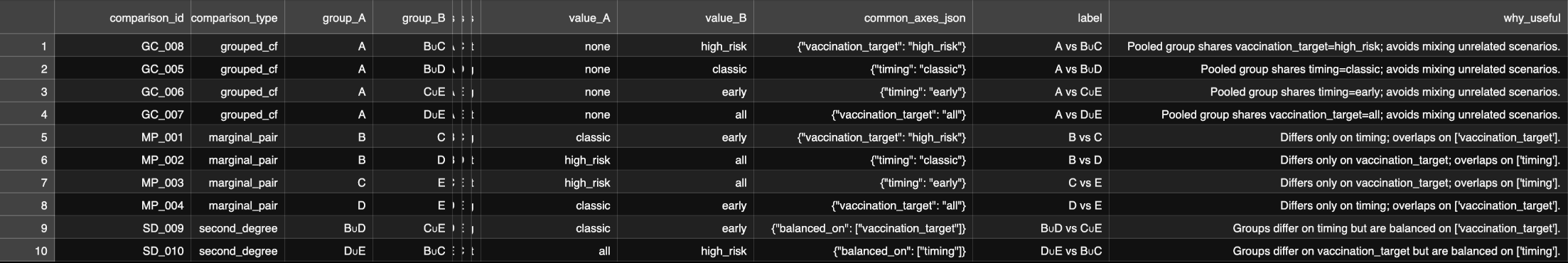}
    \caption{Excerpt from the filtered comparison table $\mathcal{C}$ produced by the comparison grammar (Algorithm~\ref{alg:grammar}). Each row is a structurally valid comparison: MP~=~marginal pair (Rule~R1, exactly one axis differs), GC~=~grouped counterfactual (Rule~R2, baseline vs.\ pooled intervention group), SD~=~second-degree (Rule~R3, balanced pool vs.\ pool). The \texttt{varying\_axis} column identifies the factor being compared; \texttt{pct\_change} columns report the relative difference between groups~A and~B for each outcome and age group.}
    \label{fig:filtered}
\end{figure}

\paragraph{Filtering with grammar rules.}
The comparison grammar (Algorithm~\ref{alg:grammar}) traverses $\mathcal{D}_{\mathrm{aug}}$ and emits only structurally valid comparisons. Rule~R1 isolates marginal effects by requiring exactly one axis of difference. Rule~R2 compares the counterfactual against a pooled intervention group, with the guard that within-group variation must be confined to modifier axes---without this constraint, the pool would mix fundamentally different interventions (e.g.\ vaccinating high-risk only vs.\ all ages), making causal attribution impossible. Rule~R3 permits pool-vs-pool comparisons only when the two sides are balanced on all non-varying axes. Figure~\ref{fig:filtered} shows the resulting filtered table: each row is a grammar-valid comparison annotated with its type, the varying axis, and the percentage change in each outcome.

\section{Pipeline}

\textbf{Algorithm~\ref{alg:style}} extracts Key Takeaways from prior-round GT reports and prompts $\mathcal{M}$ to summarize their recurring structural and stylistic patterns into a style description $S$.

\textbf{Algorithm~\ref{alg:decomp}} parses the round README for scenario metadata and calls $\mathcal{M}$ to decompose scenarios into axes $\mathcal{A}$ with primary/modifier roles $\rho$ and per-scenario values $\mathbf{t}_s$. A structural fallback (underscore splitting) is used if the LLM is unavailable.

\textbf{Algorithm~\ref{alg:schema}} maps ensemble data columns to canonical names via an identity seed refined by $\mathcal{M}$. On failure, the traceback $\epsilon^{(t)}$ is sent back to $\mathcal{M}$ for diagnosis; the corrected mapping $\sigma^{(t+1)} \leftarrow \mathcal{M}(\sigma^{(t)}, \epsilon^{(t)})$ is retried up to $T{=}4$ times.

\textbf{Algorithm~\ref{alg:augment}} enumerates all $2^{|\mathcal{S}|}{-}1$ scenario subsets, computes common/union axes and per-target statistics (min, max, mean of member medians) for each (age group, target) pair.

\textbf{Algorithm~\ref{alg:grammar}} generates candidate facts $\mathcal{C}$ via three rules: R1 (marginal pairs differing on one axis), R2 (counterfactual vs.\ pooled group, guarded against mixing primary axes), R3 (balanced pool-vs-pool). Each comparison is expanded across age groups with attached $\Delta\%$.

\textbf{Algorithm~\ref{alg:maxent}} fits a MaxEnt baseline $F$ via IPF from a uniform prior.
% scores cells with multinomial $z$-scores $z_{ij} = (M_{ij} - F_{ij})/\sqrt{F_{ij}(1-p_{ij})}$, and selects the top-$K$ with a per-field cap of $\lfloor K/2 \rfloor$.

Surprise is then scored as a multinomial z-score:
$$\small z_{ij} = \frac{M_{ij} - F_{ij}}{\sqrt{F_{ij}(1 - p_{ij})}},
\qquad p_{ij} = \frac{F_{ij}}{\sum_j F_{ij}}$$

\textbf{Algorithm~\ref{alg:summary_gen}} prompts $\mathcal{M}$ with $(S, P, \mathcal{C}^*, \mathcal{D}_{\mathrm{aug}})$ requiring a \texttt{<cot>} block per bullet. The clean summary $\hat{\mathcal{C}}$ is verified via $M_1$ (fidelity), $M_2$ (coverage), and $M_3$ (style similarity).
\subsection{Step 5: Summary Generation and Verification}
\label{app:summary}
 
\paragraph{Chain-of-thought enforcement.} The generation prompt (Algorithm~\ref{alg:summary_gen}) requires a \texttt{<cot>} block before every bullet, with four mandatory fields: \texttt{ROWS} (comparison IDs cited), \texttt{VALUES} (exact pct-change values from those rows), \texttt{CALC} (any arithmetic), and \texttt{CHECK} (confirmation that every number in the upcoming bullet appears above). After generation, \texttt{<cot>} blocks are stripped to produce the clean summary; the raw output with \texttt{<cot>} blocks is saved separately as an audit log.

\noindent
\textbf{Row-coverage check.} After generation, a regex pass extracts all comparison IDs cited in \texttt{<cot> ROWS:} lines and computes coverage = $|\text{cited} \cap \text{all\_ids}| / |\text{all\_ids}|$. Missing IDs are reported.
 
\paragraph{M1 (Factual Fidelity).} An LLM verifier checks every number in the summary against three sources: (1) the \texttt{<cot>} audit log, (2) the source comparison table, and (3) the GT Key Takeaways. Flexibility is allowed for rounding ($5\% \approx 4.8\%$), sign conventions ($-16.6\%$ reported as ``16.6\% reduction''), and simple derived arithmetic (e.g., $208{,}683 - 130{,}196 = 78{,}487$).
 
\paragraph{M2 (GT Coverage).} The evaluator extracts scenario-based GT findings (skipping trajectory descriptions and caveats), labels each as COVERED, PARTIAL, or MISSING.
 
\paragraph{M3 (Style Consistency).} Computed deterministically using TF-IDF cosine similarity (unigrams + bigrams) and optionally sentence-embedding cosine similarity (\texttt{all-MiniLM-L6-v2}). Additional diagnostics include vocabulary Jaccard overlap, average sentence length, and numeric density (fraction of sentences containing a number).

\subsection{M1 Verification: Detailed Examples}

\begin{figure*}[h]
    \centering
    \includegraphics[width=\linewidth]{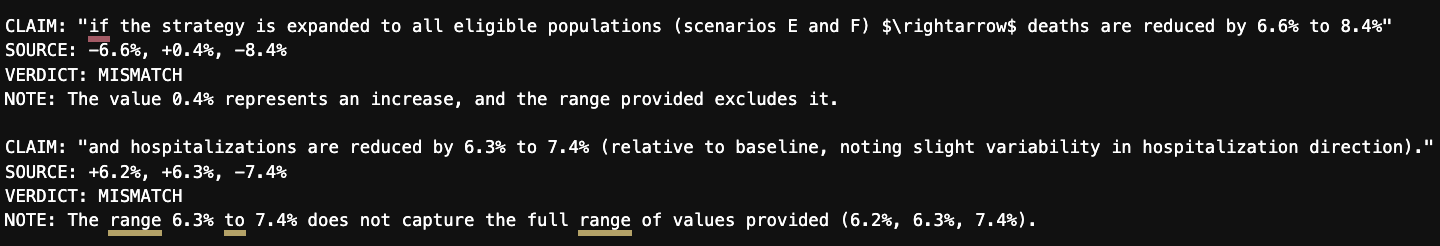}
    \caption{Example M1 (factual fidelity) verification output showing two flagged claims from a generated summary. In both cases the verifier marks the claim as \texttt{MISMATCH}: the first because the reported range excludes a positive value present in the source data, and the second because the stated range does not capture all three source values. These examples illustrate how derived statistics (ranges, averages) can introduce numerical discrepancies even when the underlying point estimates are correct.}
    \label{fig:M1_issue}
\end{figure*}

Figure~\ref{fig:M1_issue} in the main text shows two \texttt{MISMATCH} examples from the M1 verifier. We provide additional context:
 
\textbf{Example 1: Range exclusion.}
The generated summary claims ``deaths are reduced by 6.6\% to 8.4\%.'' The source data contains three values: $-6.6\%$, $+0.4\%$, $-8.4\%$. The value $+0.4\%$ represents an \emph{increase}, not a reduction, so the stated range $[6.6\%, 8.4\%]$ is misleading---it excludes a data point that reverses the direction.
 
\textbf{Example 2: Incomplete range.}
The summary states ``hospitalizations are reduced by 6.3\% to 7.4\%.'' The source values are $+6.2\%$, $+6.3\%$, $-7.4\%$. The range $[6.3\%, 7.4\%]$ omits $6.2\%$ and conflates increases with decreases (the $-7.4\%$ is a reduction; the $+6.2\%$ and $+6.3\%$ are increases).
 
These examples illustrate a systematic failure mode: when derived statistics (ranges, averages) are computed over values with mixed signs, the resulting claim can be technically ``close'' to the data but semantically incorrect.

\subsection{Algorithm}
\label{ssec:alg}
\begin{algorithm}[ht]
\caption{Step 0: Style Extraction}
\label{alg:style}
\begin{algorithmic}[1]
\Require Previous-round GT reports $\{G_r\}_{r \neq r^*}$
\Ensure Style description $S$
\For{each round $r \neq r^*$}
    \State $g_r \gets \textsc{ExtractKeyTakeaways}(G_r)$
\EndFor
\State $S \gets \mathcal{M}\!\bigl(\textsc{StylePrompt},\; \{g_r\}\bigr)$
\end{algorithmic}
\end{algorithm}

\begin{algorithm}[ht]
\caption{Step 1a: Axis Decomposition (Partial Order)}
\label{alg:decomp}
\begin{algorithmic}[1]
\Require README text $R$, flag \texttt{fallback}
\Ensure Decomposition $P = (\mathcal{A},\; \rho,\; \{\mathbf{t}_s\}_{s\in\mathcal{S}})$
    \Comment{axes, roles, per-scenario axis values}
\State $\mathit{rows} \gets \textsc{ParseScenarioTable}(R)$
\State $\mathit{desc} \gets \textsc{ParseDescriptions}(R)$
\If{\textbf{not} \texttt{fallback}}
    \State $P \gets \mathcal{M}\!\bigl(\textsc{DecomposePrompt},\; \mathit{rows},\; \mathit{desc}\bigr)$
    \If{$P$ valid and complete} \Return $P$ \EndIf
\EndIf
\State $P \gets \textsc{FallbackDecompose}(\mathit{rows})$
    \Comment{split names on `\_', all roles $\gets$ primary}
\end{algorithmic}
\end{algorithm}

\begin{algorithm}[ht]
\caption{Step 1b: Agentic Schema Inference}
\label{alg:schema}
\begin{algorithmic}[1]
\Require Parquet file $D_{\mathrm{raw}}$
\Ensure Preprocessing state $\sigma$ with validated column mapping
\State $\sigma^{(0)} \gets D_{\mathrm{raw}} \cup \mathcal{M}\!\bigl(\textsc{SchemaPrompt},\; \textsc{Describe}(D_{\mathrm{raw}})\bigr)$
\For{$t = 1, \dots, T$}
    \State Apply $\sigma^{(t-1)}$ to $D_{\mathrm{raw}}$: rename, cast, filter, aggregate
    \If{success} \Return $\sigma^{(t-1)}$
    \Else
        \State $\sigma^{(t)} \gets \mathcal{M}\!\bigl(\sigma^{(t-1)},\; \epsilon^{(t-1)}\bigr)$
            \Comment{LLM diagnoses traceback $\epsilon$}
    \EndIf
\EndFor
\end{algorithmic}
\end{algorithm}

%% ====================================================================
%% Step 2
%% ====================================================================
\begin{algorithm}[ht]
\caption{Step 2: Dataset Augmentation (Union Table)}
\label{alg:augment}
\begin{algorithmic}[1]
\Require $P$, preprocessed data $D$
\Ensure Augmented dataset $D_{\mathrm{aug}}$
\State $\mathit{stats} \gets \textsc{AgentPreprocess}(D)$
    \Comment{per (scenario, age, target): median}
\For{$r = 1$ \textbf{to} $|\mathcal{S}|$}
    \For{each $C \in \binom{\mathcal{S}}{r}$}
        \State $\mathit{common} \gets \bigcap_{s \in C} \mathbf{t}_s$
            \Comment{shared axis values}
        \State $\mathit{union} \gets \bigcup_{s \in C} \mathbf{t}_s$
        \For{each age group $a$, target $q$}
            \State Compute $\min$, $\max$, $\mathrm{mean}$ of member medians
            \State Append row to $D_{\mathrm{aug}}$
        \EndFor
    \EndFor
\EndFor
\end{algorithmic}
\end{algorithm}

%% ====================================================================
%% Step 3
%% ====================================================================
\begin{algorithm}[ht]
\caption{Step 3: Fact Generation (Comparison Grammar)}
\label{alg:grammar}
\begin{algorithmic}[1]
\Require $P = (\mathcal{A}, \rho, \{\mathbf{t}_s\})$, augmented data $D_{\mathrm{aug}}$
\Ensure Candidate fact set $\mathcal{C}$
\State $\mathcal{S}_{\mathrm{cf}} \gets \{s : s\text{ is counterfactual}\}$;\quad
       $\mathcal{S}_{\neg} \gets \mathcal{S} \setminus \mathcal{S}_{\mathrm{cf}}$
\Statex
\State \textbf{// R1: Marginal Pairs}
\For{$(s_a, s_b) \in \binom{\mathcal{S}_{\neg}}{2}$}
    \If{$|\{k : \mathbf{t}_{s_a}^k \neq \mathbf{t}_{s_b}^k\}| = 1$}
        \State $\mathcal{C} \gets \mathcal{C} \cup \{\textsc{MP}(s_a, s_b)\}$
    \EndIf
\EndFor
\Statex
\State \textbf{// R2: Grouped Counterfactual}
\For{each primary axis $k$ with $\rho(k) = \texttt{primary}$}
    \For{each value $v$, group $G_v \gets \{s \in \mathcal{S}_{\neg} : \mathbf{t}_s^k = v\}$}
        \If{within-group variation only on modifier axes}
            \State $\mathcal{C} \gets \mathcal{C} \cup \{\textsc{GC}(\mathcal{S}_{\mathrm{cf}},\; G_v)\}$
        \EndIf
    \EndFor
\EndFor
\Statex
\State \textbf{// R3: Second-Degree (Pool vs Pool)}
\For{each axis $k$, each pair of values $(v_a, v_b)$}
    \State $G_a \gets \{s : \mathbf{t}_s^k = v_a\}$,\; $G_b \gets \{s : \mathbf{t}_s^k = v_b\}$
    \State $\mathit{balanced} \gets$ (other-axis distributions match)
    \State $\mathcal{C} \gets \mathcal{C} \cup \{\textsc{SD}(G_a, G_b, \mathit{balanced})\}$
\EndFor
\Statex
\State \textbf{// Attach numerics}
\For{each $c \in \mathcal{C}$, each age group $a$, target $q$}
    \State $\Delta_c^{a,q} \gets (v_B - v_A) / |v_A| \times 100$
        \Comment{\% increment/decrement from $D_{\mathrm{aug}}$}
\EndFor
\end{algorithmic}
\end{algorithm}

%% ====================================================================
%% Step 4
%% ====================================================================
\begin{algorithm}[ht]
\caption{Step 4: Interestingness Filtering (MaxEnt / IPF)}
\label{alg:maxent}
\begin{algorithmic}[1]
\Require Observation matrix $M \in \mathbb{R}^{N \times P}$ (states $\times$ fields)
\Ensure Selected subset $\mathcal{C}^* \subseteq \mathcal{C}$
\State $\tilde{M} \gets \textsc{ZScore}(M)$
    \Comment{column-normalise}
\State $F \gets \textsc{IPF}\!\bigl(\mathbf{1}_{N\times P},\;
       \Sigma_{\mathrm{region}},\; \Sigma_{\mathrm{category}},\; \Sigma_{\mathrm{metric}}\bigr)$
    \Comment{uniform prior $\to$ marginals}
\For{each cell $(i, j)$}
    \State $p_{ij} \gets F_{ij}\, /\, \sum_j F_{ij}$
    \State $z_{ij} \gets \dfrac{M_{ij} - F_{ij}}{\sqrt{F_{ij}\,(1 - p_{ij})}}$
        \Comment{multinomial $z$-score}
\EndFor
\State $\mathcal{C}^* \gets \textsc{BalancedTopK}(\{|z_{ij}|\},\; K)$
    \Comment{cap $\lfloor K/2 \rfloor$ per field}
\State $H_K \gets -\sum_i \hat{p}_i \log \hat{p}_i$
    \Comment{selection entropy}
\end{algorithmic}
\end{algorithm}

\begin{algorithm}[ht]
\caption{Step 5: Summary Generation \& Verification}
\label{alg:summary_gen}
\begin{algorithmic}[1]
\Require $S$, $P$, $\mathcal{C}^*$, $D_{\mathrm{aug}}$, scenario descriptions
\Ensure Summary $\hat{C}$, verification scores $(M_1, M_2, M_3)$
\Statex
\State \textbf{// 5a: Generate}
\State $\mathit{prompt} \gets (S,\; P,\; \mathcal{C}^*,\; D_{\mathrm{aug}},\; \mathit{checklist})$
\State $\hat{C}_{\mathrm{raw}} \gets \mathcal{M}(\mathit{prompt})$
    \Comment{with mandatory \texttt{<cot>} per bullet}
\State $\hat{C} \gets \textsc{StripCoT}(\hat{C}_{\mathrm{raw}})$
\Statex
\State \textbf{// 5b: Verify}
\State $M_1 \gets N_{\mathrm{verified}} \;/\; N_{\mathrm{claims}}$
    \Comment{factual fidelity via LLM}
\State $M_2 \gets (N_{\mathrm{covered}} + \tfrac{1}{2}N_{\mathrm{partial}}) \;/\; N_{\mathrm{GT}}$
    \Comment{GT coverage}
\State $M_3 \gets \cos(e_{\hat{C}},\; e_{\mathrm{GT}})$
    \Comment{style similarity}
\end{algorithmic}
\end{algorithm}

\begin{figure}[h]
    \centering
    \includegraphics[width=\linewidth]{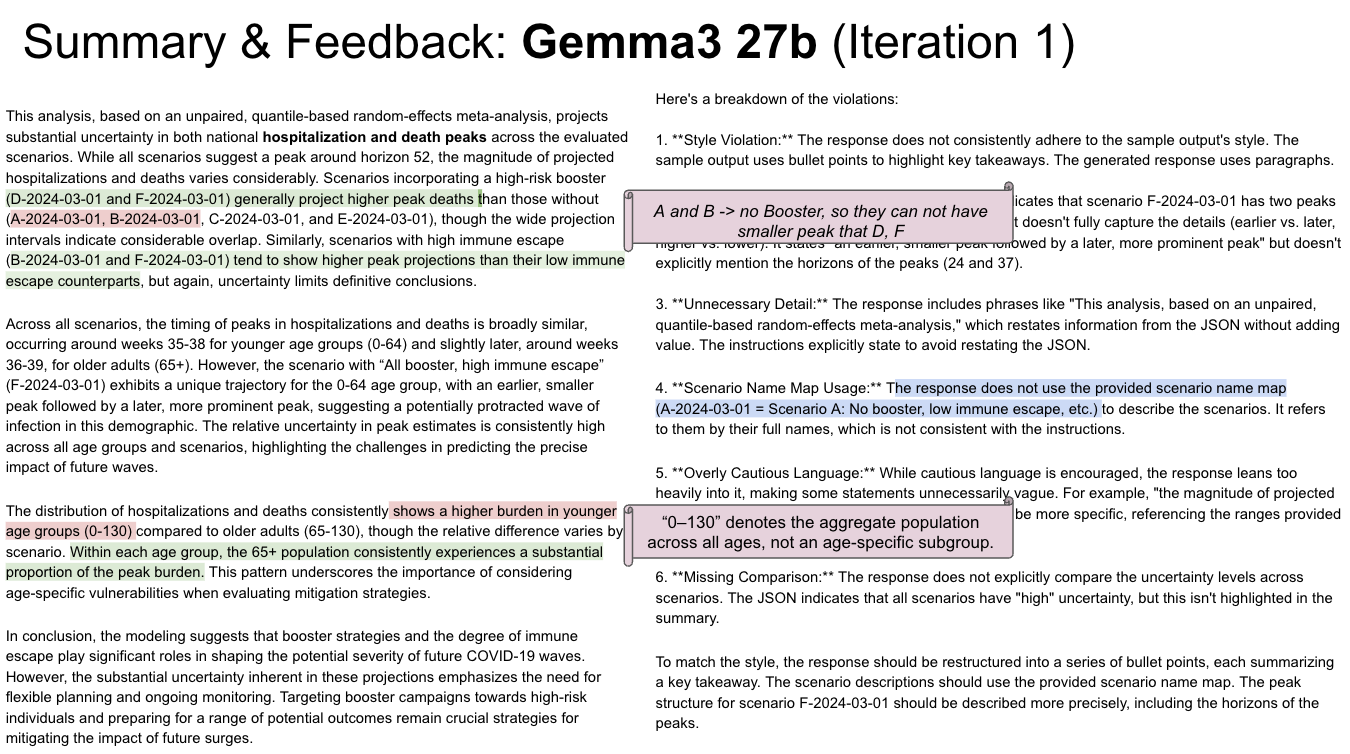}
    \caption{ Iterative summary refinement. Left: the initial Gemma~3 27B output after one round of generation, showing the raw summary, and the annotations highlight specific issues: missing scenario names, inaccurate claims, and wrong age-group comparison. Right: feedback on the summary, however, the verification alone can't identify the mistakes, hence we need the structured representation and cot.
}
    \label{fig:placeholder}
\end{figure}

\section{Case Study: Round 19 Scenario Design}
\begin{figure}
    \centering
    \includegraphics[width=0.9\linewidth]{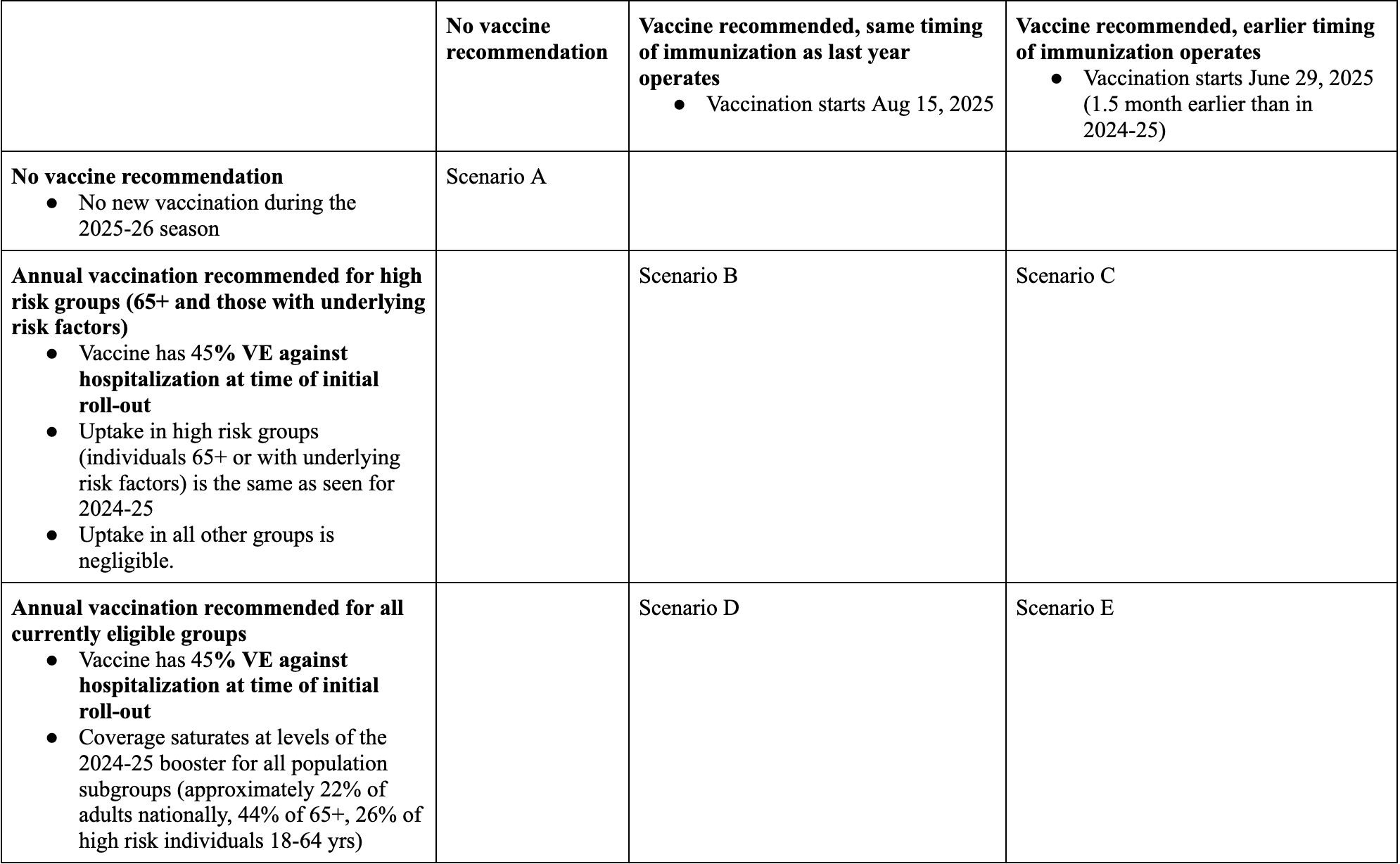}
    \caption{Round 19 Scenarios}
    \label{fig:sc19}
\end{figure}

Round~19 of the COVID-19 Scenario Modeling Hub evaluates the impact of annual booster vaccination strategies for the 2025--26 season. Five scenarios (A--E) are defined along two axes (Figure~\ref{fig:sc19}):

\begin{itemize}
    \item \textbf{Booster target} (primary axis): \texttt{none} (Scenario~A, the no-vaccination counterfactual), \texttt{high\_risk} (adults 65+ and immunocompromised; Scenarios~B, C), or \texttt{all\_eligible} (all currently eligible age groups; Scenarios~D, E).
    \item \textbf{Immunization timing} (modifier axis): \texttt{classic} (vaccination campaign beginning mid-August, consistent with prior seasons; Scenarios~B, D) or \texttt{early} (campaign shifted earlier, starting late June; Scenarios~C, E).
\end{itemize}
 
This yields a $2 \times 2$ factorial over the intervention scenarios plus a shared counterfactual, enabling three types of comparison via the grammar. The four marginal pairs (e.g.\ B~vs.~C, isolating the effect of timing within the high-risk target) each differ on exactly one axis. Two grouped-counterfactual comparisons pool over timing (a modifier axis) to estimate the overall effect of each target population against the baseline. One second-degree comparison contrasts the high-risk pool \{B,\,C\} against the all-eligible pool \{D,\,E\}, balanced on timing. The decomposition agent classifies \texttt{booster\_target} as primary and \texttt{immunization\_timing} as modifier, which activates the R2 guard: pooling over timing is permitted (it is a modifier), but pooling over target population is blocked in grouped-CF comparisons (it is primary). Figure \ref{fig:augmented} and \ref{fig:filtered} show a partial image of the augmented table and the filtered entry, respectively.
 
\subsection{Chain-of-Thought Narrative: Role and Utility}
\label{app:cot}
 
\begin{figure}[h]
    \centering
    \includegraphics[width=\linewidth]{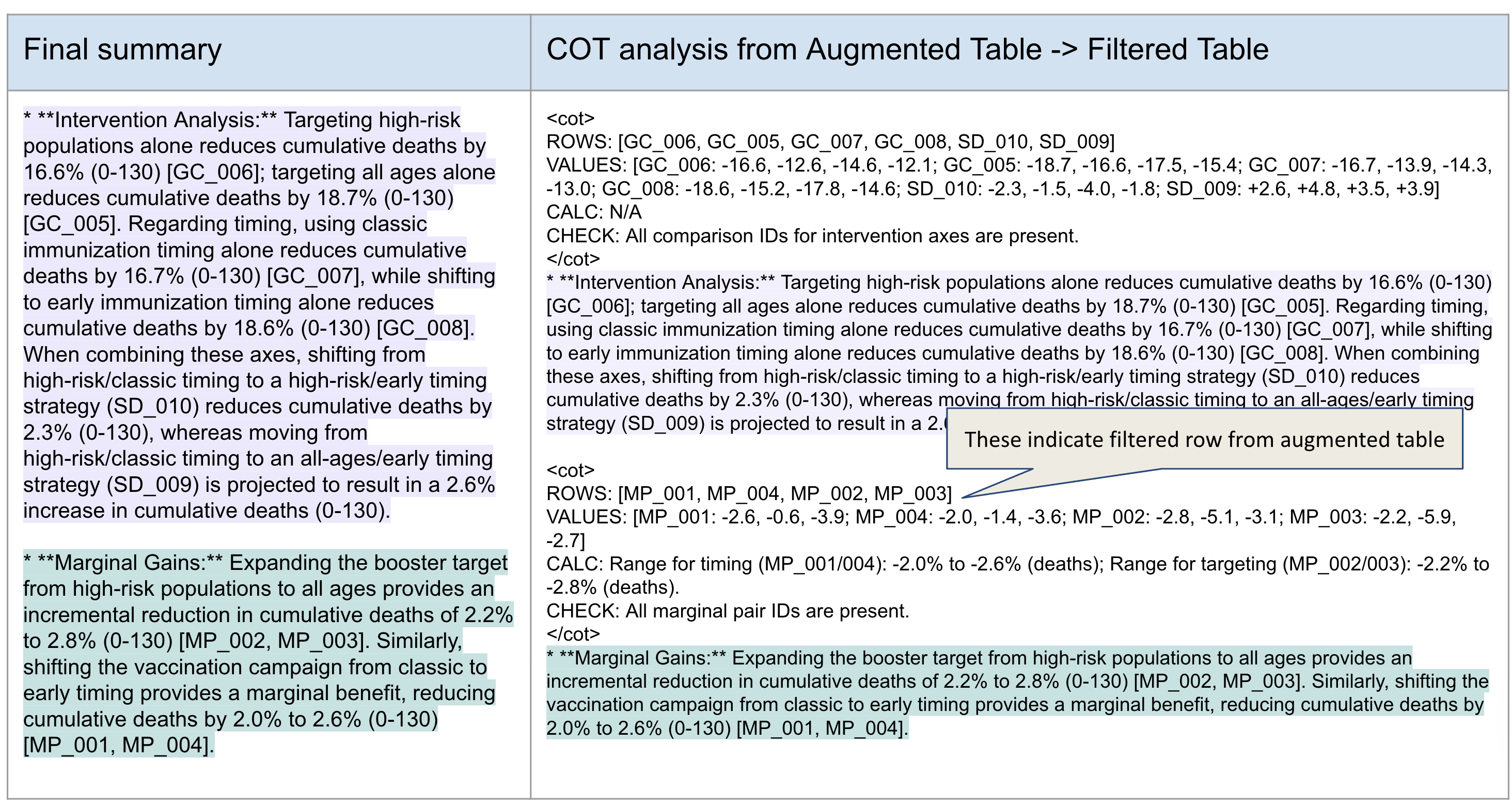}
    \caption{Chain-of-thought audit trail for the final \tool{} summary (Round~19). Left column: the clean summary bullets as they appear in the output. Right column: the corresponding \texttt{<cot>} blocks showing the comparison IDs cited (\texttt{ROWS}), the exact percentage-change values pulled from the filtered table (\texttt{VALUES}), any arithmetic performed (\texttt{CALC}), and the consistency check (\texttt{CHECK}). Highlighted spans show where filtered-table row IDs map to specific claims in the final text, demonstrating end-to-end traceability from augmented data to narrative.
}
    \label{fig:cot_final}
\end{figure}

A common failure mode of LLM-generated summaries is \emph{untraceable arithmetic}: the model reports a number that cannot be verified against any source row, either because it was hallucinated or because the derivation path is opaque. The mandatory \texttt{<cot>} block addresses this by requiring the model to show its work before every bullet (Figure~\ref{fig:cot_final}).
 
Each \texttt{<cot>} block contains four fields:
\begin{enumerate}
    \item \texttt{ROWS}: the comparison IDs (e.g.\ \texttt{GC\_005}, \texttt{MP\_001}) from the filtered table (Table \ref{fig:filtered}) $\mathcal{C}$ that the upcoming bullet draws on.
    \item \texttt{VALUES}: the exact percentage-change values copied from those rows (e.g.\ $-16.6\%$, $-14.3\%$).
    \item \texttt{CALC}: any arithmetic applied---ranges, means, differences, or absolute-count derivations (e.g.\ $208{,}683 - 130{,}196 = 78{,}487$ averted hospitalizations).
    \item \texttt{CHECK}: a self-verification line confirming that every number in the upcoming bullet appears in \texttt{VALUES} or \texttt{CALC}.
\end{enumerate}
 
This structure provides three concrete benefits:

\noindent
\textbf{1. Post-hoc auditability.} The cot output creates a full provenance chain from augmented data to narrative. A reviewer, either human or automated, can trace any claim backward: bullet $\to$ \texttt{<cot>} $\to$ comparison ID $\to$ row in $\mathcal{D}_{\mathrm{aug}}$ $\to$ source parquet. The M1 verifier exploits this chain directly, checking numbers against the \texttt{VALUES} and \texttt{CALC} fields before falling back to the raw source tables.

\noindent
\textbf{Reduced hallucination.} Requiring the model to cite specific row IDs \emph{before} writing the bullet constrains generation to claims that are grounded in the filtered table. In the ablation (Table~\ref{tab:ablation_study}), removing CoT drops M1 from 100\% to 70\% while increasing the claim count from 30 to 46---the model produces more numbers but with weaker grounding, because it is no longer forced to anchor each claim to a specific comparison.

\noindent
\textbf{Coverage enforcement.} The row checklist in the prompt lists every comparison ID that must appear in at least one \texttt{ROWS} field. After generation, a regex pass computes the fraction of IDs cited. Missing IDs are reported, and if coverage is below a threshold, the prompt can be reissued with explicit instructions to address the gaps. This mechanism ensures that the summary does not silently omit important comparisons.

\subsection{Additional MaxEnt-based Interestingness Analysis}
\label{sec:maxent_results}

To identify state-level patterns that are not apparent from aggregate national summaries alone, we apply a maximum-entropy interestingness analysis to the state-level projection outputs. The MaxEnt baseline is fitted using iterative proportional fitting (IPF), simultaneously matching HHS regional marginals (row axis) and field-category/metric marginals (column axis).
All absolute burden and averted-count fields are expressed per 100{,}000 population using 2022 ACS five-year estimates before the matrix is constructed, so that state size does not mechanically inflate residual
scores.
Large absolute residuals indicate state--metric cells that are surprising
relative to what would be expected from the marginal structure alone.

\noindent
\textbf{1. State-level deviation.}
After population normalization, the most surprising states overall are
concentrated in \textbf{HHS Region~1 (New England)}: Vermont
($\bar{z}=3.48$, $\max=11.25$), Massachusetts ($\bar{z}=3.29$),
New Hampshire ($\bar{z}=3.23$), Maine ($\bar{z}=3.20$), and Rhode
Island ($\bar{z}=3.07$).
Within this cluster, Vermont, New Hampshire, and Maine are primarily
driven by higher-than-expected rates of averted hospitalizations under
Scenario~B (classic timing, high-risk booster) relative to their
regional baseline, with Vermont reaching $74.2$ averted hospitalizations
per 100{,}000 population---well above the MaxEnt expectation.
Vermont and New Hampshire additionally show a higher-than-expected
incremental gain when expanding from Scenario~B to Scenario~D
(all-eligible booster, same classic timing), at $30.9$ and $22.9$ per
100{,}000 respectively, suggesting that broadening booster eligibility
yields disproportionate returns in these states.

West Virginia ($\bar{z}=3.04$, $\max=11.63$) stands out within
Region~3 (Mid-Atlantic) as a high-burden outlier, with a projected
median hospitalization rate of $281.2$ per 100{,}000 under Scenario~B,
significantly exceeding regional expectations.
Kentucky ($220.5$/100k) and New York ($246.8$/100k) also show
higher-than-expected hospitalization burden rates.
Conversely, the District of Columbia ($5.4$/100k), Rhode Island
($10.9$/100k), and Utah ($5.9$/100k) exhibit projected death rates that
are notably \emph{below} the MaxEnt expectation, representing a
distinct cluster of lower-than-anticipated mortality burden.

\emph{Evaluation and justification.}
The selected findings reflect deviations that cannot be explained by
regional or metric-level marginals alone.
Following per-capita normalization, findings are no longer driven by
state population size; instead, they reflect genuine rate-based
heterogeneity across intervention efficacy, hospitalization burden, and
mortality concentration.
The New England booster-responsiveness cluster and the West Virginia
high-burden outlier represent the two primary dimensions of surprise:
intervention-related deviations and absolute burden deviations.
The lower-than-expected death rates in DC, Rhode Island, and Utah
constitute a third dimension warranting further investigation.

\noindent
\textbf{2. Age-specific mortality concentration.}
Alaska (86\%), New Mexico (85\%), and Mississippi (87\%) exhibit
higher-than-expected shares of projected deaths occurring in the
65$+$ age group, all above the national historical baseline of
approximately 75--80\% for COVID-19 mortality, despite the 65$+$
population representing only $\sim$17\% of the US population.
These findings are based on relative share measures and are independent
of state population size.
Rhode Island shows a complementary signal: a lower-than-expected share
of \emph{hospitalizations} in the 0--64 age group (38\%), consistent
with a more elderly-concentrated hospitalization burden.
Complementary age-share fields (e.g., high 65$+$ share and low 0--64
share for the same state and metric) are deduplicated in the selection
procedure so that each represents a distinct finding.

\noindent
\textbf{3. MaxEnt selection diagnostics.}
To assess whether the selected findings are overly concentrated in one
type of signal, we examine the category composition of the top-15
MaxEnt cells and the per-field cap applied during selection (at most
$\lfloor N/4 \rfloor = 3$ cells per field for $N=15$).
The selected cells span four categories: age-share
($\bar{|z|}=5.00$, 4 cells), burden ($\bar{|z|}=4.58$, 4 cells),
absolute averted ($\bar{|z|}=2.13$, 3 cells), and incremental averted
($\bar{|z|}=1.34$, 2 cells).
This distribution reflects the category-level surprise scores: age-share
and burden quantities show the largest mean deviations from the MaxEnt
baseline, while percentage-reduction quantities ($\bar{|z|}=2.56$) and
incremental effects are comparatively more consistent with the fitted
marginals.
At the regional level, Region~1 (New England) exhibits the highest mean
surprise ($3.19$), followed by Region~5 (Chicago, $2.91$) and
Region~3 (Mid-Atlantic, $2.90$), pointing to a broader clustering of
surprising patterns across the Northeast and Midwest.

Overall, the MaxEnt analysis provides a complementary view of the ensemble projections by identifying where state-level patterns deviate most strongly from the expected marginal structure.

Rather than replacing the national summary, this analysis helps prioritize which state--metric combinations may deserve closer inspection, particularly in the context of resource allocation and targeted public health intervention.

\begin{figure}
    \centering
    \includegraphics[width=0.5\linewidth]{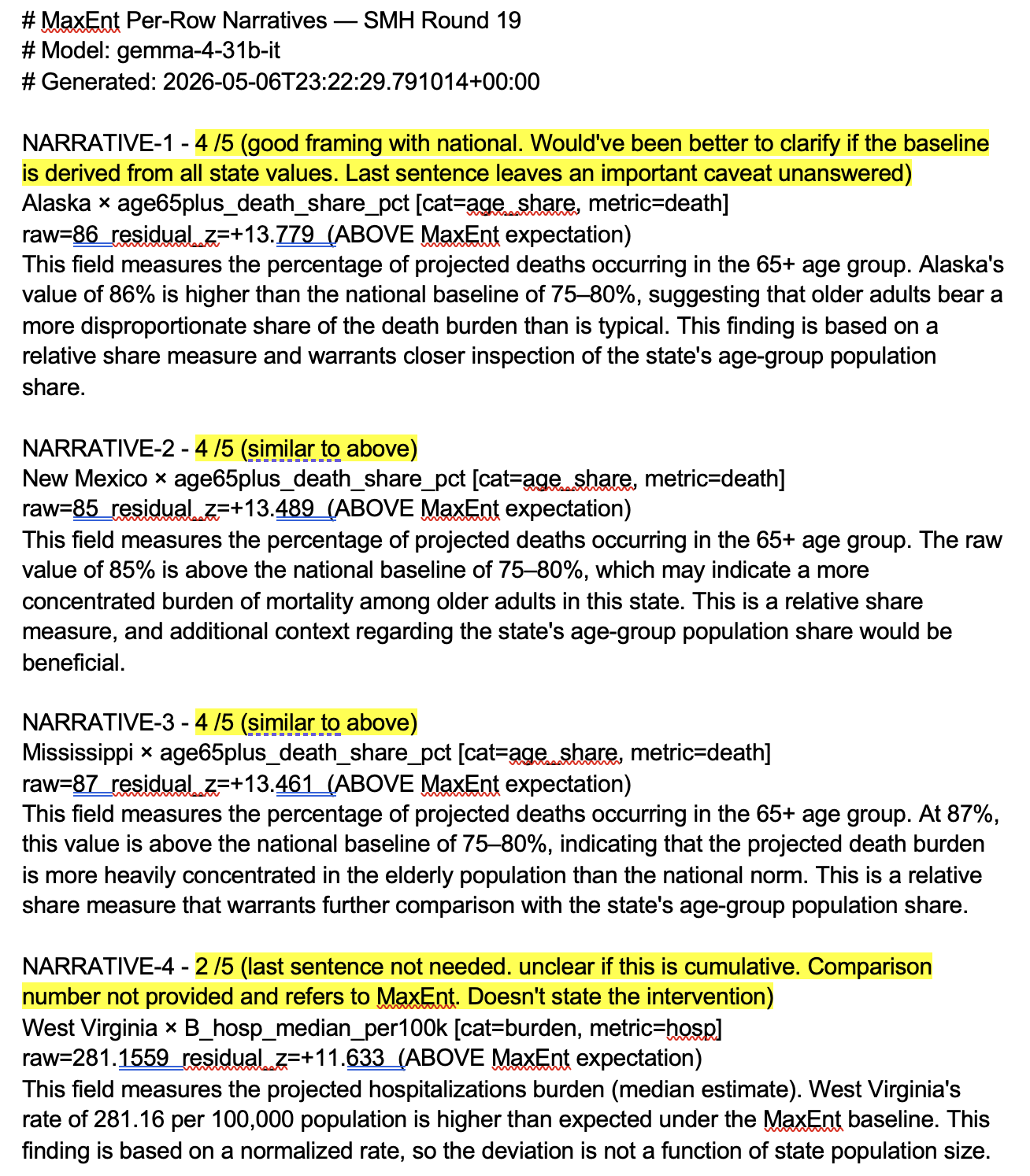}
\caption{Expert evaluation of the MaxEnt results (1)}
\label{fig:maxent1}
\end{figure}

\begin{figure}
    \centering
    \includegraphics[width=0.5\linewidth]{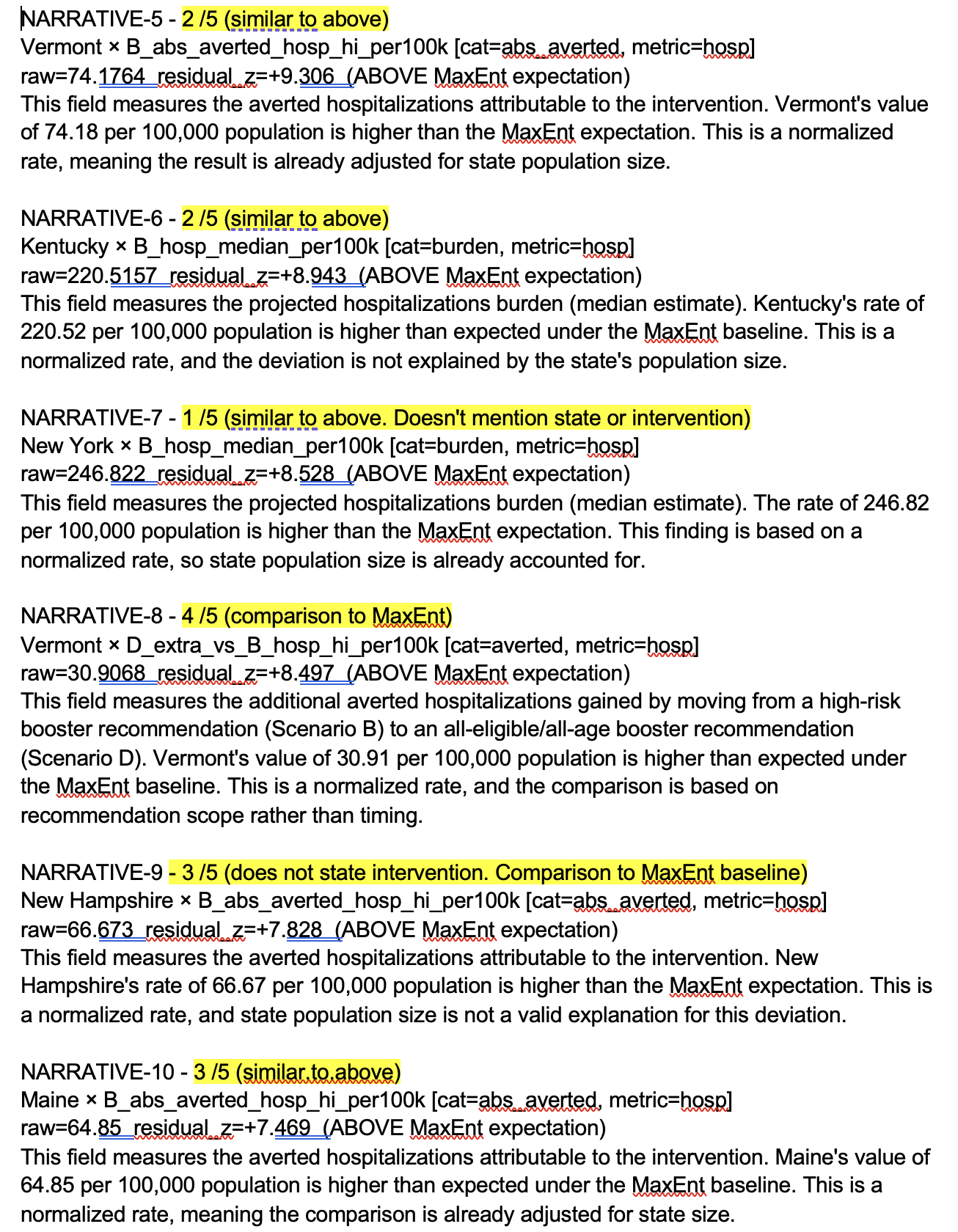}
\caption{Expert evaluation of the MaxEnt results (2)}
\label{fig:maxent2}
\end{figure}

\begin{figure}
    \centering
    \includegraphics[width=0.5\linewidth]{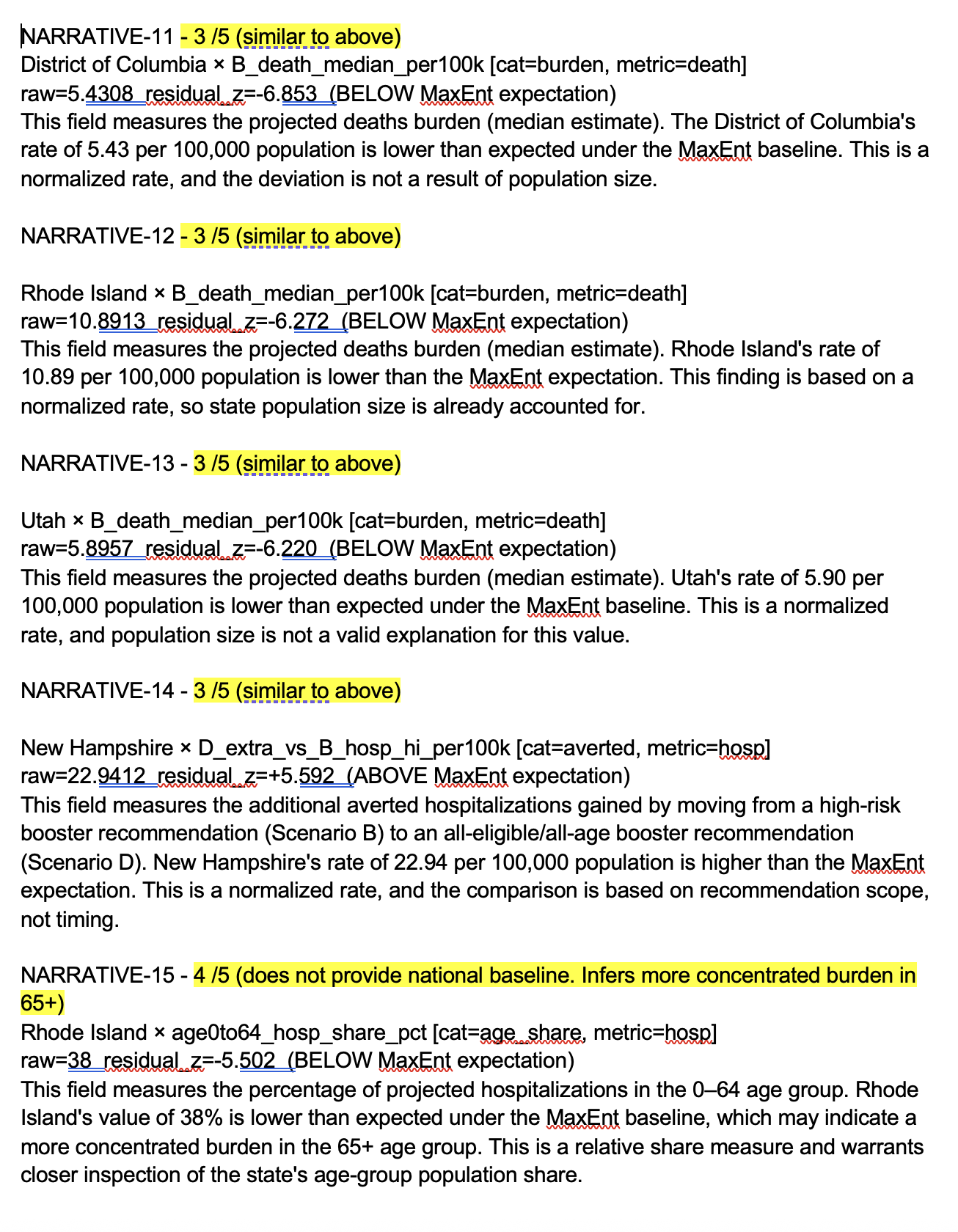}
\caption{Expert evaluation of the MaxEnt results (3)}
\label{fig:maxent3}
\end{figure}

We further evaluate the selected statements with an epidemiology expert, who scores each statement based on its narrative interestingness. This evaluation addresses the question: \emph{Are the top-$K$ MaxEnt-selected factors actually interesting to domain experts?} We find that the interestingness-based selection adds value to the final narrative, since experts prioritize state-level and regional burden patterns as well as scenario-wise comparative facts that may otherwise be missed in aggregate summaries. Figures~\ref{fig:maxent1}, \ref{fig:maxent2}, and \ref{fig:maxent3} show expert ratings for the top 15 MaxEnt-selected facts. Across these facts, epidemiology experts rated 33\% as interesting enough to retain in the final narrative.

\begin{figure}
    \centering
    \includegraphics[width=\linewidth]{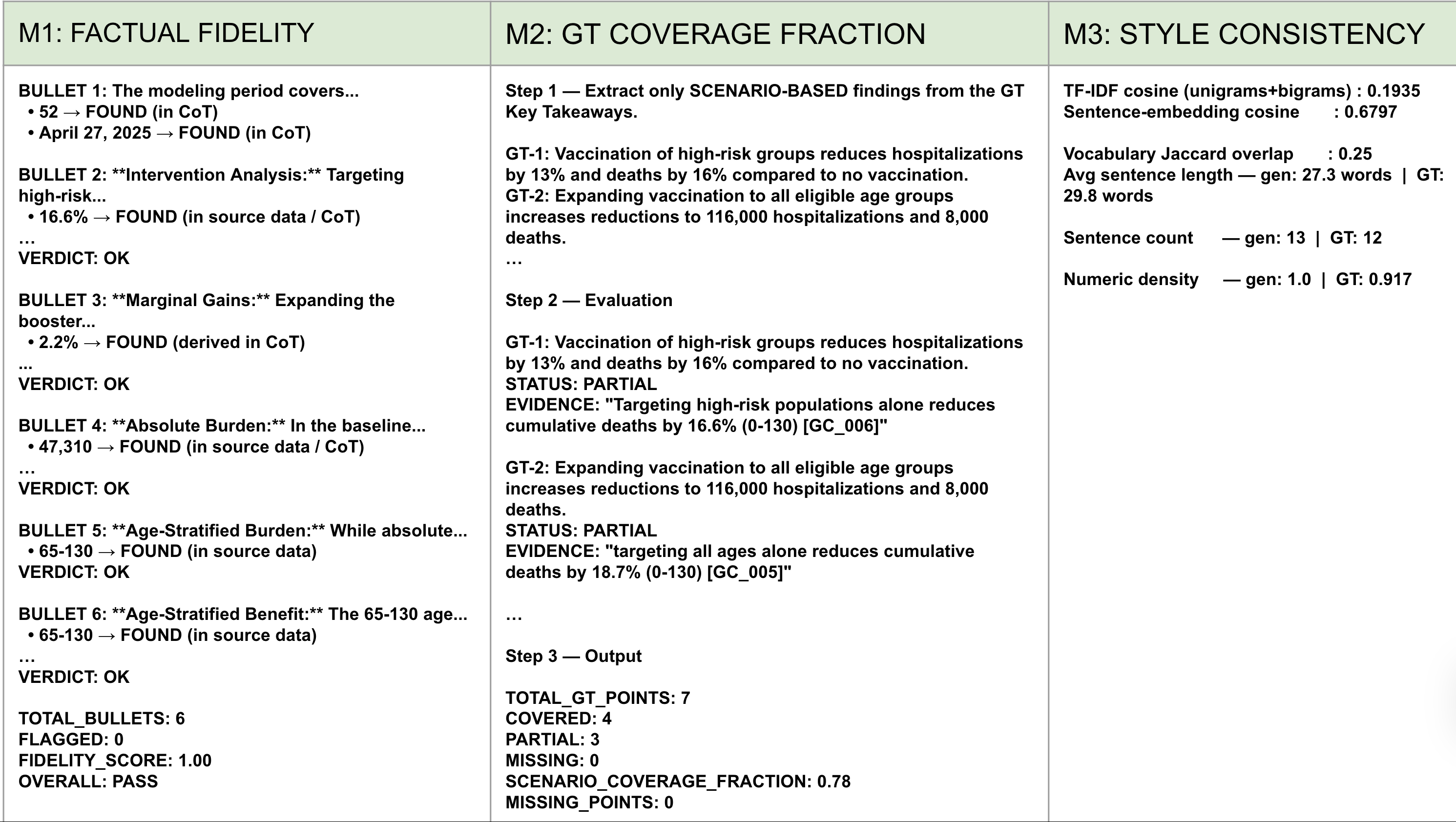}
    \caption{Round 19 evaluation on \tool{}}
    \label{fig:metrics}
\end{figure}

% \newpage
% \input{checklist.tex}

\end{document}